\documentclass[a4paper,fleqn]{cas-dc}

\usepackage[numbers]{natbib}

\usepackage{algorithmic}
\usepackage{graphicx}
\usepackage{algorithm,algorithmic}
\usepackage{hyperref}
\usepackage[capitalize]{cleveref}
\usepackage{makecell}
\usepackage{booktabs}  
\usepackage{multibib}

\usepackage{mwe} 
\usepackage{array, xcolor, colortbl}
\usepackage{multirow}
\usepackage{booktabs}
\usepackage{upgreek}
\usepackage{mathtools}
\usepackage{dsfont}
\usepackage{hyperref}

\begin{document}
\let\WriteBookmarks\relax
\def\floatpagepagefraction{1}
\def\textpagefraction{.001}

\shorttitle{Leveraging Foundation Models for Content-Based Image Retrieval in Radiology}    

\shortauthors{Denner et al.}  

\title [mode = title]{Leveraging Foundation Models for Content-Based Image Retrieval in Radiology}

\author[1,2]{Stefan Denner}
\cormark[1]
\ead{stefan.denner@dkfz-heidelberg.de}
\credit{Writing - original draft, Visualization, Validation, Software, Methodology, Formal analysis, Data curation, Conceptualization, Project administration}

\author[1]{David Zimmerer}
\credit{Writing – review \& editing, Conceptualization, Methodology}

\author[1,3]{Dimitrios Bounias}
\credit{Writing – review \& editing, Conceptualization, Methodology}

\author[1,3]{Markus Bujotzek}
\credit{Writing – review \& editing, Conceptualization, Methodology}

\author[1,2]{Shuhan Xiao}
\credit{Writing – review \& editing, Formal analysis}

\author[1,2]{Raphael Stock}
\credit{Writing – review \& editing, Formal analysis}

\author[1]{Lisa Kausch}
\credit{Writing – review \& editing, Methodology}

\author[1,2]{Philipp Schader}
\credit{Writing – review \& editing, Methodology}

\author[4]{Tobias Penzkofer}
\credit{Writing – review \& editing, Methodology, Formal analysis}

\author[5]{Paul F. Jäger}
\credit{Writing – review \& editing, Conceptualization, Methodology}

\author[1,2,3]{Klaus Maier-Hein}
\credit{Writing – review \& editing, Supervision, Funding acquisition, Resources, Conceptualization}

\affiliation[1]{organization={German Cancer Research Center (DKFZ) Heidelberg, Division of Medical Image Computing},
            city={Heidelberg},
            country={Germany}}

\affiliation[2]{organization={Faculty of Mathematics and Computer Science, Heidelberg University},
            city={Heidelberg},
            country={Germany}}

\affiliation[3]{organization={Medical Faculty Heidelberg, University of Heidelberg},
            city={Heidelberg},
            country={Germany}}

\affiliation[4]{organization={Department of Radiology, Charité - Universitätsmedizin Berlin},
            city={Berlin},
            country={Germany}}

\affiliation[5]{organization={German Cancer Research Center (DKFZ) Heidelberg, Interactive Machine Learning Group},
            city={Heidelberg},
            country={Germany}}

\cortext[1]{Corresponding author}

\begin{abstract}
Content-based image retrieval (CBIR) has the potential to significantly improve diagnostic aid and medical research in radiology.
However, current CBIR systems face limitations due to their specialization to certain pathologies, limiting their utility.
On the other hand, several vision foundation models have been shown to produce general-purpose visual features. 
Therefore, in this work, we propose using vision foundation models as powerful and versatile off-the-shelf feature extractors for content-based image retrieval.
Our  contributions include: (1) benchmarking a diverse set of vision foundation models on an extensive dataset comprising 1.6 million 2D radiological images across four modalities and 161 pathologies; (2) identifying weakly-supervised models, particularly BiomedCLIP, as highly effective, achieving a achieving a P@1 of up to 0.594 (P@3: 0.590, P@5: 0.588, P@10: 0.583), comparable to specialized CBIR systems but without additional training; (3) conducting an in-depth analysis of the impact of index size on retrieval performance; (4) evaluating the quality of embedding spaces generated by different models; and (5) investigating specific challenges associated with retrieving anatomical versus pathological structures.
Despite these challenges, our research underscores the vast potential of foundation models for CBIR in radiology, proposing a shift towards versatile, general-purpose medical image retrieval systems that do not require specific tuning. 
Our code, dataset splits and embeddings are publicly available 
\href{https://github.com/MIC-DKFZ/foundation-models-for-cbmir}{here}.

\end{abstract}



\maketitle

\section{Introduction}

\begin{figure*}[t]
    \centering
    \includegraphics[width=0.9\textwidth]{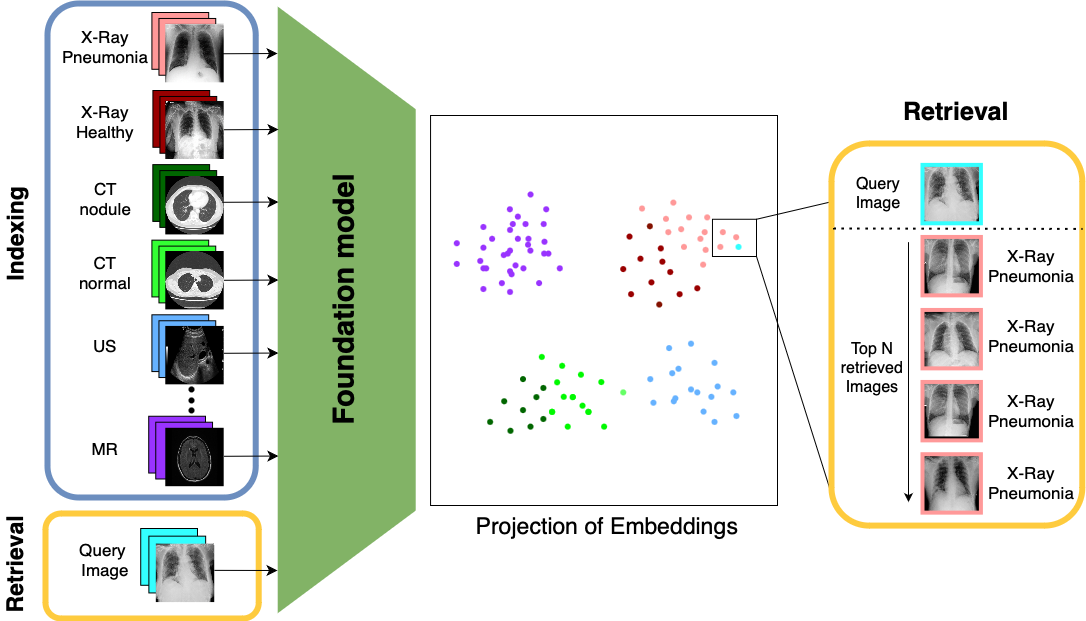}
    \caption{
        Visualization of the content-based image retrieval system workflow using foundation models.
        During indexing, we use an off-the-shelf foundation model to generate embeddings for the medical images and store them in a vector database. 
        During retrieval, we compare the embeddings of the query image to those in the index and retrieve the most similar ones.
    }
    \label{fig:fig1}
\end{figure*}

Medical imaging is a cornerstone of modern healthcare, playing a critical role in diagnosing, monitoring, and managing a wide range of diseases \cite{bushberg2011essential,puttagunta2021medical,PreparingMedicalImagingDataForML,pham2000current}.
With the growing adoption of advanced imaging modalities such as X-ray (XR), computed tomography (CT), magnetic resonance imaging (MRI), and ultrasound (US), radiology has seen an exponential increase in the volume and diversity of image data generated daily \cite{mazurowski2019deep,bushberg2011essential}. 
This abundance of imaging data presents both an opportunity and a challenge: while it provides a rich source of information for clinical decision-making and research, effectively managing, retrieving, and utilizing these data remains a formidable task \cite{PreparingMedicalImagingDataForML,3dMIR,denner2023efficient}.

Content-Based Image Retrieval (CBIR) systems address this challenge by enabling efficient and accurate retrieval of relevant images based on their visual and contextual features rather than relying solely on metadata or textual descriptions \cite{muller2004review,akgul2011content}.
To achieve this, CBIR systems store low-dimensional representations of images in a database and retrieve similar images based on representation similarity \cite{muller2004review}.

In recent years, a key advancement in these systems has been the adoption of deep learning models to encode the visual features of images \cite{DeepImageRetrieval,cvnet}. 
To that end, a neural network is trained using a contrastive loss, encouraging the feature vectors of similar images (positive pairs) to be close together in the feature space, while pushing the feature vectors of dissimilar images (negative pairs) further apart \cite{DeepImageRetrieval,cvnet}. 
Despite these advancements, existing CBIR systems in the medical domain remain constrained, as they are usually trained on a limited set of pathologies, hindering their ability to generalize to unseen conditions \cite{ClinicalMIR,XMIR,regionbasedMIR,SketchbasedMIR,ozturk2021attention,ozturk2023content}.
This is in stark contrast to the evolving needs of radiology, which demand a more general and versatile medical image retrieval system capable of handling a broader scope of pathologies \cite{futureOfRadiology,muller2004review,decadeMIRSurvey,denner2023efficient}.

Developing a feature extractor capable of accurately retrieving a wide variety of pathologies presents several challenges.
Radiological datasets are highly heterogeneous, encompassing diverse imaging modalities, anatomical variations, and subtle pathological differences, often with limited or imbalanced annotations \cite{PreparingMedicalImagingDataForML,akgul2011content}. Strict privacy regulations and fragmented access to comprehensive datasets further hinder the training of general-purpose models \cite{kondylakis2023data}. 
Additionally, current CBIR systems trained on specific pathologies struggle to generalize to unseen conditions (\cref{sec:contextualize_performance}). 
These issues are compounded by the semantic gap between low-level visual features and high-level clinical understanding, making it difficult to design systems that align with the needs of radiological practice.
These limitations underscore the need for alternative approaches to general-purpose CBIR in radiology.  
Recently, vision foundation models have shown remarkable performance on a variety of tasks by not directly optimizing for the target task (e.g. semantic segmentation, object classification), but rather performing a large-scale pretraining resulting in all-purpose robust visual features \cite{CLIP,Dinov2}.
We hypothesize that due to their exposure to a diverse range of images, vision foundation models possess a nuanced understanding of relevant features, including those of modality, anatomy, and pathology \cite{BiomedCLIP,MedSAM,MedCLIP,3dMIR,bayerMIR}.
Therefore, in this work, we investigate the use of vision foundation models as powerful and versatile off-the-shelf feature extractors for content-based image retrieval in radiology.

Our contributions include:
\begin{itemize}
    \item Aggregating and curating multiple publicly available radiological 2D datasets, spanning in total 1.6 million images across four modalities (CT, MR, X-ray, and US), and 185 classes. This combined dataset is one of the most comprehensive in the field and is designed to support the evaluation of CBIR systems in radiology.
    \item Assessing the performance of various vision foundation models, differentiating in their training scheme, pretraining dataset and parameters.
    \item Provide a detailed performance analysis comparing off-the-shelf vision foundation models to specialized models, trained on the combined dataset, showcasing that those specialized models are still superior. 
    \item Investigate the impact of the index size on the retrieval performance of medical CBIR systems, offering insights into the required data support for image retrieval performance.
    \item Quantitatively analyzing the embedding space of the vision foundation models to assess how well retrieval-relevant features are captured in the embedding space. 
\end{itemize}

To support future research, all resources are made publicly available, including dataset splits, embeddings, weights and code.

The remainder of this paper is organized as follows. Section 2 describes the datasets utilized and details the evaluated foundation models. It further outlines the retrieval pipeline, performance metrics, and benchmarking methodology. Section 3 presents the experimental results, analyzing the retrieval performance of foundation models across modalities, comparing anatomical and pathological retrieval effectiveness, and providing qualitative visualizations of embeddings and retrieval examples. Section 4 discusses the implications of these findings, highlighting strengths, limitations, and potential future research directions. Finally, Section 5 summarizes the key insights and conclusions drawn from this study.

\section{Materials and Methods}

\subsection{Datasets}
To comprehensively evaluate a foundation model's performance in content-based image retrieval in radiology, we combine four public datasets -- RadImageNet \cite{RadImageNet}, NIH14 \cite{NIH14}, MIMIC \cite{MIMIC} and CheXpert \cite{CheXpert} -- resulting in a comprehensive collection of over 1.6 million 2D images spanning four modalities, 12 anatomical regions, and 185 classes (161 pathological, 24 anatomical).

\subsubsection*{NIH14}
The NIH14 dataset, also known as the ChestX-ray14 dataset, is a large-scale collection of chest X-ray images provided by the National Institutes of Health (NIH). It contains 112,120 frontal-view X-ray images of 30,805 unique patients, with annotations for 14 different thoracic disease labels \cite{NIH14}. 

\subsubsection*{MIMIC}
The MIMIC-CXR dataset is part of the Medical Information Mart for Intensive Care (MIMIC) project, which includes a vast repository of de-identified medical records. MIMIC-CXR focuses on chest X-rays, providing a large dataset of 377,110 images from 227,827 radiographic studies, collected from 65,379 patients. Structured labels (12 thoracic disease labels) for these images were extracted from the associated free-text radiology reports \cite{MIMIC}.

\subsubsection*{CheXpert}
CheXpert is a large-scale dataset of chest radiographs, consisting of 224,316 images from 65,240 patients, annotated with labels for 14 different pathologies. The dataset includes labels that are generated using a rule-based labeler, which classifies findings into positive, negative, and uncertain categories based on the information extracted from radiology reports \cite{CheXpert}.

\subsubsection*{RadImageNet}
The RadImageNet database consists of 1.35 million annotated medical images in 131,872 patients who underwent CT, MRI, and US for musculoskeletal, neurologic, oncologic, gastrointestinal, endocrine, abdominal, and pulmonary pathologic conditions.
The dataset contains 165 labels, with 155 disease and 10 anatomical labels. 

\subsubsection*{Combined Dataset}
Radiology is inherently multi-modal, therefore, we combine the four datasets. 
Due to overlapping labels in the X-ray datasets, we harmonize them, ensuring that the same pathology across different datasets is assigned a consistent label.
Furthermore, we exclude multi-label samples in all X-ray datasets to maintain a single-label focus. 
The RadImageNet dataset is already single label.
We split all datasets according to their provided patient-wise dataset splits, utilizing the train set for creating the index and the test set for querying (\cref{tab:datasets_retrieval}).

Our combined dataset enables a highly realistic evaluation of retrieval systems by reflecting the inherent long-tailed distribution of medical images. 
In clinical practice, a small subset of diseases is frequently encountered, while the majority are rare and underrepresented \cite{longtail}. 
The combined dataset mirrors this distribution, exhibiting a significant class imbalance where the smallest class contains only 13 samples, and the largest exceeds 226,000 samples (Appendix \cref{fig:imbalance}).

\begin{table}[h]
    \centering
    \caption{
        Datasets utilized for the retrieval benchmark. Classes are formatted as \{\#pathological\}+\{\#anatomical\} classes.
    }
    \label{tab:datasets_retrieval}
    \begin{tabular}{c|c|cc|c}
        \textbf{Dataset}               & \textbf{Mod.} & \textbf{Index} & \textbf{Query} &  \textbf{Classes} \\ 
        \midrule
        NIH14            & XR                & 66,192               & 17,853        & 14+1                              \\ 
        MIMIC           & XR                & 229,597         & 2,473         & 12+1                              \\
        CheXpert      & XR                & 50,194             & 54            & 12+1                              \\ 
        RadImageNet & MR                & 460,479          & 57,103        & 109+7                             \\
        \multicolumn{1}{l|}{}          & CT                & 202,521             & 63,198        & 32+2                              \\
        \multicolumn{1}{l|}{}          & US                & 313,569           & 31,072        & 1+14                              \\ 
        \midrule
        \multirow{2}{*}{\shortstack{Combined \\ Dataset}}                          &                   & 1,322,552       & 171,753       & 161+24
    \end{tabular}
\end{table}

\begin{table*}[!t]
    \caption{
        Foundation models utilized for the benchmark.
        We evaluate a range of foundation models on their capabilities as a feature extractor for content-based image retrieval in radiology. We evaluate across training scheme (supervised, weakly-, self-supervised), pretraining datasets (natural, medical) and architectures.
    }
    \centering
    \label{tab:models}
    \begin{tabular}{cccccccc}
        \toprule
        \textbf{Method}              & \textbf{Architecture} & \textbf{Domain} & \textbf{\#Params.} & \textbf{\#Features} &  \textbf{Input Size}   & \makecell{\textbf{\#Training} \\ \textbf{Images}} & \makecell{\textbf{Pretraining} \\ \textbf{Datasets}}            \\ 
        \midrule
        \multicolumn{8}{c}{\textbf{Supervised}}          \\
        \midrule
        
        ResNet \cite{ResNet}         & ResNet50              & Natural         & 25 M        & 1000  & $224^2$       & 1.3 M             & IN1K \cite{Imagenet}         \\
        ViT \cite{ViT}               & ViT-B/16              & Natural         & 91 M      & 768  & $224^2$          & 14 M              & IN21K \cite{Imagenet}        \\
        Ark \cite{ark} & Swin-B & CXR & 88 M & 1024 & $224^2$ & 704 K & Multiple CXR \cite{ark} \\
        SAM   \cite{SAM}             & ViT-B/16              & Natural         & 91 M     & 768   & $1024^2$           & 11 M             & SA-1B  \cite{SAM}                      \\
        MedSAM \cite{MedSAM}         & ViT-B/16              & Medical         & 91 M   & 768 & $1024^2$             & 1.6 M             & custom \cite{MedSAM}               \\
        \midrule
        \multicolumn{8}{c}{\textbf{Weakly-Supervised}} \\ 
        \midrule
        CLIP \cite{CLIP}             & ViT-L/14              & Natural         & 300 M      & 1024  & $224^2$        & 400 M             & WIT-400M    \cite{CLIP}                 \\
        
        MedCLIP \cite{MedCLIP}      & Swin-T                & CXR         & 29 M      & 768  & $224^2$         & 600 K             & \makecell{MIMIC \cite{MIMIC} \& \\  CheXpert \cite{CheXpert}} \\
        BiomedCLIP \cite{BiomedCLIP} & ViT-B/16              & Biomedical      & 91 M       & 768  & $224^2$        & 15 M              & PMC-15M   \cite{BiomedCLIP}                   \\ 
        BMC-CLIP \cite{BIOMEDICA} & ViT-L/14 & Biomedical & 300 M & 768 & $224^2$ & 24 M & BIOMEDICA \cite{BIOMEDICA}
                \\ 

        \midrule
        
        \multicolumn{8}{c}{\textbf{Self-Supervised}} \\ 
        \midrule
        MAE  \cite{MAE}              & ViT-B/16              & Natural         & 91 M      & 768    & $224^2$         & 14 M              & IN21K \cite{Imagenet}        \\
        DINOv2 \cite{Dinov2}         & ViT-B/14              & Natural         & 87 M   & 768    & $224^2$            & 142 M             & LVD-142M \cite{Dinov2} \\
        RAD-DINO \cite{raddino} & ViT-B/14 & CXR & 87 M & 768 & $518^2$ & 838 K & Multi-CXR \cite{raddino} \\
        \bottomrule

    \end{tabular}
\end{table*}

\subsection{Foundation models}
\label{sec:fm}
We choose a range of foundation models spanning self-supervised, weakly-supervised, and fully-supervised learning schemes, trained on both medical and natural images, ensuring a broad evaluation spectrum (\cref{tab:models}).

\subsubsection*{Supervised models}
Our baselines are ResNet~\cite{ResNet} and Vision Transformer (ViT) \cite{ViT}, which are supervised classification models trained on ImageNet \cite{Imagenet}. 
We use ResNet's probability class predictions as feature embeddings, for ViT we use the class token.

As a medical supervised classification foundation model, we add Ark \cite{ark}. It is specifically designed to accrue and reuse diagnostic knowledge from heterogeneous expert annotations across numerous public chest X-ray datasets. Empirical results demonstrate that Ark consistently outperforms state-of-the-art fully supervised, self-supervised, and domain-adapted models across a range of classification and segmentation tasks. Furthermore, Ark exhibits superior embedding quality for linear probing and enhanced robustness to underdiagnosis and demographic bias.
We use the embeddings of the teacher network.

Recently, supervised prompt-based segmentation foundation models such as the Segment Anything Model (SAM) \cite{SAM} demonstrated remarkable performance on a wide variety of segmentation tasks, including zero-shot. 
SAM was trained on a dataset of more than 300 million natural images with 1 billion masks.
While it is not trained with any explicit semantic supervision, experimental evidence suggests that the model captures semantic information in its representations \cite{SAM}, making it a compelling candidate for inclusion in this benchmark.

Additionally, we include MedSAM \cite{MedSAM}, a medical adaptation of SAM. 
MedSAM was trained on a dataset of more than 1.5 million image-mask pairs spanning 10 imaging modalities and over 30 pathologies. 
It has demonstrated superior accuracy and robustness than modality-wise specialist models. 
These results suggest that MedSAM could serve as an effective general-purpose feature extractor across diverse medical imaging tasks.

For both models, SAM and MedSAM, we utilize the average patch tokens of the image encoder as embeddings. 

\subsubsection*{Weakly-Supervised models}
For weakly-supervised models, we focus on CLIP-based approaches due to their ability to align the representations of images and text in a shared embedding space. CLIP \cite{CLIP} was trained on large-scale data to understand visual concepts and their relationships in a more abstract and generalized manner than traditional supervised models, which are trained for specific tasks. This results in CLIP's exceptional zero-shot capabilities in natural image domains, making it highly versatile for unseen tasks and a strong candidate for inclusion as a feature extractor in our benchmark.

MedCLIP~\cite{MedCLIP}, a medical adaptation of CLIP, is specifically trained on Chest X-ray images and has demonstrated superior performance in zero-shot prediction and supervised classification for radiological tasks. While it remains to be empirically validated, there is potential for the knowledge MedCLIP gains from X-ray pathologies to be transferable to other medical imaging modalities.

BiomedCLIP~\cite{BiomedCLIP}, a biomedical adaptation of CLIP, is trained on 15 million biomedical image-text pairs collected from 4.4 million scientific articles. The high fidelity and detailed descriptions of figures in these articles suggest that BiomedCLIP has developed a nuanced and pronounced understanding of biomedical concepts. This rich dataset enables BiomedCLIP to capture fine-grained semantic relationships, leading to its superior performance over state-of-the-art radiology-specific models in radiology tasks. Given its training on diverse and high-quality scientific content, BiomedCLIP is particularly well-suited as a high-performing, general-purpose feature extractor for a wide array of biomedical tasks.

BMC-CLIP \cite{BIOMEDICA} extends this concept even further, leveraging a similar training paradigm but expanding the scale and diversity of its training data. Specifically, it was trained on over 24 million unique biomedical image-text pairs drawn from more than 6 million articles. This allowed BMC-CLIP to demonstrate robust generalization across biomedical domains. On average, it achieves state-of-the-art performance across 40 biomedical tasks covering diverse fields such as radiology, pathology, ophthalmology, dermatology, surgery, molecular biology, parasitology, and cell biology.

For all CLIP-based models, we use the vision encoder representations as embeddings.

\subsubsection*{Self-Supervised models}
Self-supervised approaches have emerged as highly effective methods for learning general-purpose visual representations without relying on large labeled datasets. By learning from the data itself, these models capture rich and transferable features that are not limited to the specific tasks they were trained on. This makes them particularly useful in medical imaging, where annotated data can be scarce or expensive to obtain. Including self-supervised models in this benchmark allows us to evaluate how well these models generalize across a variety of medical imaging modalities and tasks, and their performance without task-specific supervision is of particular interest.

Masked Autoencoder (MAE)~\cite{MAE} employs a masked image modeling approach, where large portions of the input image are masked and the model is tasked with reconstructing the missing regions. This pretext task forces the model to capture high-level semantic information, enabling it to generalize well across different tasks. MAE's efficient, scalable architecture has proven to outperform supervised methods in a variety of downstream applications such as classification, object detection, and segmentation, making it a promising candidate for image feature extraction in this benchmark, despite being trained on natural images.

DINOv2~\cite{Dinov2} is a self-supervised approach that builds on contrastive learning techniques, focusing on capturing both image-level and patch-level information. Trained on a highly diverse and curated dataset of natural images, DINOv2 achieves superior performance across numerous vision tasks without the need for task-specific fine-tuning. Its ability to learn robust and transferable visual features, even at the pixel level, makes it particularly well-suited for medical imaging tasks where detailed structural understanding is critical.

Additionally, we include RAD-DINO~\cite{raddino}, a medical adaptation of DINOv2 specifically pretrained on chest X-ray images. RAD-DINO learns visual representations without textual supervision, demonstrating strong performance across various chest X-Ray tasks, including classification, semantic segmentation, and vision-language alignment tasks. Notably, RAD-DINO highlights the scalability and effectiveness of image-only self-supervision, achieving state-of-the-art results on multiple medical imaging benchmarks.

For all self-supervised models, we utilize the class token of the image encoder as feature embeddings.

\subsection{Specialized CBIR Model}
To contextualize the efficacy of employing vision foundation models as feature extractors, it is critical to benchmark their performance against dedicated feature extractors for content-based image retrieval.

To this end, we benchmark the foundation models against CVNet \cite{cvnet}, the current state-of-the-art training paradigm on the $\mathcal{R}$Oxford dataset \cite{oxford,revisitingOxford}. 
We train the global backbone network of CVNet (CVNet-Global) on our combined dataset using the objective loss \cite{berman2019multigrain}, which jointly optimizes the classification loss and contrastive loss to induce the network to learn more distinctive and robust global and local representations \cite{cvnet}. Training details can be found in \cref{sec:training_details_cvnet}. 
We train two architectures, namely ResNet50 and ResNet101.
We refer to those models as \textit{Specialists}.

\subsection{Leveraging foundation models for image retrieval}
\label{sec:fm4retrieval}

\begin{table*}[h!]
\centering
    \caption{Retrieval performance of the foundation models on the combined dataset. BiomedCLIP emerges as the best-performing foundation model, but is inferior to the trained specialist. }
    \label{tab:retrieval}
\begin{tabular}{c|cccc|cccc}
\toprule
                    & \multicolumn{4}{c|}{Sample-wise (micro)}                                      & \multicolumn{4}{c}{Class-wise (macro)}                                        \\ 
                    & \textbf{P@1}      & \textbf{P@3}      & \textbf{P@5}      & \textbf{P@10}     & \textbf{P@1}      & \textbf{P@3}      & \textbf{P@5}      & \textbf{P@10}     \\
\midrule
\multicolumn{9}{c}{\textbf{Off-the-shelf}} \\
\midrule
ResNet & {\cellcolor[HTML]{FFF7B2}} \color[HTML]{000000} 0.539 & {\cellcolor[HTML]{FFF8B4}} \color[HTML]{000000} 0.535 & {\cellcolor[HTML]{FFFAB6}} \color[HTML]{000000} 0.531 & {\cellcolor[HTML]{FFFCBA}} \color[HTML]{000000} 0.526 & {\cellcolor[HTML]{EEF8A8}} \color[HTML]{000000} 0.203 & {\cellcolor[HTML]{F2FAAE}} \color[HTML]{000000} 0.192 & {\cellcolor[HTML]{F5FBB2}} \color[HTML]{000000} 0.187 & {\cellcolor[HTML]{ECF7A6}} \color[HTML]{000000} 0.180 \\
ViT & {\cellcolor[HTML]{B3DF72}} \color[HTML]{000000} 0.560 & {\cellcolor[HTML]{BBE278}} \color[HTML]{000000} 0.554 & {\cellcolor[HTML]{BDE379}} \color[HTML]{000000} 0.550 & {\cellcolor[HTML]{C1E57B}} \color[HTML]{000000} 0.543 & {\cellcolor[HTML]{A5D86A}} \color[HTML]{000000} 0.217 & {\cellcolor[HTML]{B7E075}} \color[HTML]{000000} 0.204 & {\cellcolor[HTML]{B5DF74}} \color[HTML]{000000} 0.199 & {\cellcolor[HTML]{B3DF72}} \color[HTML]{000000} 0.190 \\
Ark & {\cellcolor[HTML]{128A49}} \color[HTML]{F1F1F1} 0.586 & {\cellcolor[HTML]{138C4A}} \color[HTML]{F1F1F1} 0.582 & {\cellcolor[HTML]{148E4B}} \color[HTML]{F1F1F1} 0.579 & {\cellcolor[HTML]{15904C}} \color[HTML]{F1F1F1} 0.574 & {\cellcolor[HTML]{006837}} \color[HTML]{F1F1F1} 0.245 & {\cellcolor[HTML]{006837}} \color[HTML]{F1F1F1} 0.234 & {\cellcolor[HTML]{006837}} \color[HTML]{F1F1F1} 0.228 & {\cellcolor[HTML]{006837}} \color[HTML]{F1F1F1} 0.218 \\
SAM & {\cellcolor[HTML]{FDAF62}} \color[HTML]{000000} 0.522 & {\cellcolor[HTML]{FDBF6F}} \color[HTML]{000000} 0.519 & {\cellcolor[HTML]{FDC171}} \color[HTML]{000000} 0.516 & {\cellcolor[HTML]{FECA79}} \color[HTML]{000000} 0.511 & {\cellcolor[HTML]{FEEB9D}} \color[HTML]{000000} 0.193 & {\cellcolor[HTML]{FEEFA3}} \color[HTML]{000000} 0.185 & {\cellcolor[HTML]{FEEFA3}} \color[HTML]{000000} 0.180 & {\cellcolor[HTML]{FFF5AE}} \color[HTML]{000000} 0.173 \\
MedSAM & {\cellcolor[HTML]{A50026}} \color[HTML]{F1F1F1} 0.491 & {\cellcolor[HTML]{A50026}} \color[HTML]{F1F1F1} 0.484 & {\cellcolor[HTML]{A50026}} \color[HTML]{F1F1F1} 0.479 & {\cellcolor[HTML]{A50026}} \color[HTML]{F1F1F1} 0.471 & {\cellcolor[HTML]{A50026}} \color[HTML]{F1F1F1} 0.154 & {\cellcolor[HTML]{A50026}} \color[HTML]{F1F1F1} 0.145 & {\cellcolor[HTML]{A50026}} \color[HTML]{F1F1F1} 0.141 & {\cellcolor[HTML]{A50026}} \color[HTML]{F1F1F1} 0.133 \\
\midrule
CLIP & {\cellcolor[HTML]{73C264}} \color[HTML]{000000} 0.571 & {\cellcolor[HTML]{70C164}} \color[HTML]{000000} 0.567 & {\cellcolor[HTML]{6EC064}} \color[HTML]{000000} 0.565 & {\cellcolor[HTML]{70C164}} \color[HTML]{000000} 0.559 & {\cellcolor[HTML]{87CB67}} \color[HTML]{000000} 0.222 & {\cellcolor[HTML]{7AC665}} \color[HTML]{000000} 0.213 & {\cellcolor[HTML]{75C465}} \color[HTML]{000000} 0.209 & {\cellcolor[HTML]{5AB760}} \color[HTML]{F1F1F1} 0.202 \\
MedCLIP & {\cellcolor[HTML]{C5E67E}} \color[HTML]{000000} 0.557 & {\cellcolor[HTML]{C7E77F}} \color[HTML]{000000} 0.552 & {\cellcolor[HTML]{CBE982}} \color[HTML]{000000} 0.547 & {\cellcolor[HTML]{D1EC86}} \color[HTML]{000000} 0.540 & {\cellcolor[HTML]{CFEB85}} \color[HTML]{000000} 0.210 & {\cellcolor[HTML]{D7EE8A}} \color[HTML]{000000} 0.199 & {\cellcolor[HTML]{D7EE8A}} \color[HTML]{000000} 0.194 & {\cellcolor[HTML]{D5ED88}} \color[HTML]{000000} 0.185 \\
BiomedCLIP & {\cellcolor[HTML]{006837}} \color[HTML]{F1F1F1} 0.594 & {\cellcolor[HTML]{006837}} \color[HTML]{F1F1F1} 0.590 & {\cellcolor[HTML]{006837}} \color[HTML]{F1F1F1} 0.588 & {\cellcolor[HTML]{006837}} \color[HTML]{F1F1F1} 0.583 & {\cellcolor[HTML]{0C7F43}} \color[HTML]{F1F1F1} 0.240 & {\cellcolor[HTML]{0A7B41}} \color[HTML]{F1F1F1} 0.230 & {\cellcolor[HTML]{0B7D42}} \color[HTML]{F1F1F1} 0.224 & {\cellcolor[HTML]{026C39}} \color[HTML]{F1F1F1} 0.217 \\
BMC-CLIP & {\cellcolor[HTML]{04703B}} \color[HTML]{F1F1F1} 0.592 & {\cellcolor[HTML]{026C39}} \color[HTML]{F1F1F1} 0.589 & {\cellcolor[HTML]{016A38}} \color[HTML]{F1F1F1} 0.587 & {\cellcolor[HTML]{026C39}} \color[HTML]{F1F1F1} 0.582 & {\cellcolor[HTML]{2DA155}} \color[HTML]{F1F1F1} 0.233 & {\cellcolor[HTML]{15904C}} \color[HTML]{F1F1F1} 0.226 & {\cellcolor[HTML]{108647}} \color[HTML]{F1F1F1} 0.223 & {\cellcolor[HTML]{08773F}} \color[HTML]{F1F1F1} 0.215 \\
\midrule
MAE & {\cellcolor[HTML]{6BBF64}} \color[HTML]{000000} 0.572 & {\cellcolor[HTML]{69BE63}} \color[HTML]{F1F1F1} 0.569 & {\cellcolor[HTML]{63BC62}} \color[HTML]{F1F1F1} 0.567 & {\cellcolor[HTML]{5AB760}} \color[HTML]{F1F1F1} 0.562 & {\cellcolor[HTML]{128A49}} \color[HTML]{F1F1F1} 0.238 & {\cellcolor[HTML]{0E8245}} \color[HTML]{F1F1F1} 0.229 & {\cellcolor[HTML]{118848}} \color[HTML]{F1F1F1} 0.222 & {\cellcolor[HTML]{0D8044}} \color[HTML]{F1F1F1} 0.213 \\
DINOv2 & {\cellcolor[HTML]{DAF08D}} \color[HTML]{000000} 0.552 & {\cellcolor[HTML]{D5ED88}} \color[HTML]{000000} 0.548 & {\cellcolor[HTML]{D7EE8A}} \color[HTML]{000000} 0.545 & {\cellcolor[HTML]{D5ED88}} \color[HTML]{000000} 0.539 & {\cellcolor[HTML]{A5D86A}} \color[HTML]{000000} 0.217 & {\cellcolor[HTML]{B1DE71}} \color[HTML]{000000} 0.205 & {\cellcolor[HTML]{B5DF74}} \color[HTML]{000000} 0.199 & {\cellcolor[HTML]{ADDC6F}} \color[HTML]{000000} 0.191 \\
RAD-DINO & {\cellcolor[HTML]{9DD569}} \color[HTML]{000000} 0.564 & {\cellcolor[HTML]{9DD569}} \color[HTML]{000000} 0.560 & {\cellcolor[HTML]{9BD469}} \color[HTML]{000000} 0.557 & {\cellcolor[HTML]{98D368}} \color[HTML]{000000} 0.552 & {\cellcolor[HTML]{6BBF64}} \color[HTML]{000000} 0.226 & {\cellcolor[HTML]{7AC665}} \color[HTML]{000000} 0.213 & {\cellcolor[HTML]{82C966}} \color[HTML]{000000} 0.207 & {\cellcolor[HTML]{78C565}} \color[HTML]{000000} 0.198 \\
\midrule
\multicolumn{9}{c}{\textbf{Specialist}}          \\
\midrule
CVNetGlobal50 & 0.644 & 0.643 & 0.642 & 0.642 & 0.251 & 0.244 & 0.238 & 0.232 \\
CVNetGlobal101 & 0.650 & 0.649 & 0.649 & 0.647 & 0.266 & 0.258 & 0.252 & 0.245 \\
\bottomrule
\end{tabular}
\end{table*}

To evaluate the effectiveness and versatility of vision foundation models as feature encoders for content-based image retrieval (CBIR) in radiology, we adopt an off-the-shelf approach. 
This allows us to assess the models’ inherent capacity to extract meaningful and generalizable image representations without task-specific fine-tuning. 
The retrieval pipeline is structured as follows:
\paragraph{Preprocessing} Each image is resized to the input dimensions required by the foundation model (see \cref{tab:models} for input sizes). This ensures compatibility while retaining sufficient visual detail for feature extraction.
\paragraph{Feature Extraction} 

Each image is passed through a vision foundation model to obtain a dense embedding that encodes semantic features, ideally relevant to radiology, such as modality, anatomical context, and pathology. 
These embeddings form the basis for similarity comparisons in the retrieval task.
Formally, given an image $I \in \mathbb{R}^{H \times W \times C}$, where $H$, $W$, and $C$ denote height, width, and number of channels, respectively, the foundation model defines a parameterized mapping
$$\Phi_\theta: \mathbb{R}^{H \times W \times C} \longrightarrow \mathbb{R}^D,$$
that transforms an input image $I$ into a $D$-dimensional embedding $z=\Phi_\theta(I)$ (see \cref{tab:models} for embedding sizes). 
\paragraph{Normalization} 

To standardize the scale of embeddings and enable meaningful cosine similarity comparisons, we normalize each feature vector to unit length. 
This ensures consistent comparison across all embeddings, independent of their original magnitudes.
Let $z\in\mathbb{R}^D$ be the raw embedding. We apply $L_2$ normalization:
\[
\tilde z = \frac{z}{\|z\|_2}
\quad\text{where}\quad
\|z\|_2 = \sqrt{\sum_{i=1}^D z_i^2}
\]
This normalization places all vectors on the unit hypersphere.

\paragraph{Indexing} The normalized embeddings are stored in a vector database using FAISS \cite{FAISS}, which provides efficient and scalable nearest-neighbor search capabilities. This indexing step is critical for enabling real-time retrieval over large datasets.
Let $\tilde z_i$ denote the normalized embedding of the $i$-th indexed image.
\paragraph{Retrieval} 
For a query image $I_q$, its normalized embedding 
$\tilde z_q$ is computed as above. 
We then score every database vector $\tilde z_i$ against $\tilde z_q$ using the cosine similarity. 
Since our vectors are normalized, cosine similarity becomes the dot product with
$ s_i = \tilde z_q^\top \tilde z_i$.

The top $N$ images with the highest similarity scores $s_i$ are retrieved and presented in descending order of similarity \cite{manning2009introduction}.

This workflow highlights the simplicity and scalability of leveraging foundation models in a CBIR setting. By utilizing off-the-shelf models, we eliminate the need for expensive data labeling and fine-tuning.

\subsection{Evaluation}
We measure the retrieval performance between the query image and the top N retrieved images, quantified by Precision at N ($P@N$) \cite{manning2009introduction}, defined as: 
\begin{equation}
    \label{eq:p_at_n}
     P @ N=\frac{1}{QN} \sum_{q=1}^Q \sum_{j=1}^N rel_q(j),
\end{equation}

where $rel_q(j) \in \{0,1\}$ is the relevance of the $j$-th image for the $q$-th query, $Q$ is the number of queries and $N \in \{1, 3, 5, 10\}$.
In our setting, an image is considered relevant if it belongs to the same class as the query image.
To gain a more comprehensive understanding of the model's performance, especially for classes with fewer examples, we calculate both micro- and macro-averaged scores.
Micro-averaging treats all samples equally, regardless of their class membership. It measures the average score across all samples (See \cref{eq:p_at_n}).
Macro-averaging computes $P@N$ independently for each class and then averages across all classes. Let $C$ be the number of classes and $Q_c$ be the number of queries from class $c$. Then:
\begin{equation}
    \text{macro }P@N = \frac{1}{C} \sum_{c=1}^{C} \left( \frac{1}{Q_c N} \sum_{q=1}^{Q_c} \sum_{j=1}^{N} \text{rel}_q(j) \right)
\end{equation}
Macro-averaging prevents dominant classes from skewing the overall score and provides a more balanced view of model performance across both frequent and rare categories.

\section{Results}

\subsection{Retrieval performance of foundation models} \label{sec:retrieval}
In our evaluation of foundation models on the combined dataset (\cref{tab:retrieval}), BiomedCLIP emerges as the top performer, achieving a P@1 of 0.594. BMC-CLIP follows closely with a comparable P@1 score of 0.592.
The strong performance of these two models, both trained on large-scale biomedical image-text datasets, underscores the effectiveness of extensive scientific image-text alignment in enhancing medical image retrieval by capturing semantically rich and nuanced representations.

Classification supervised models, ResNet and ViT, perform moderately well despite their training on natural image datasets. 
ViT consistently outperforms ResNet, likely benefiting from its transformer-based architecture, which captures richer contextual information. 

Ark, although exclusively trained on chest X-ray images, demonstrates considerable generalizability beyond its original modality. It ranks third in micro P@1 and notably achieves the highest macro P@1, reflecting its effective transferability to underrepresented classes.

However, all supervised models lag behind biomedical CLIP-based models, which can probably be attributed to the latter’s broader training on image-text pairs, fostering more transferable representations.

Segmentation-based approaches show generally low performance, likely because such models predominantly learn to recognize structural connectivity rather than semantic meaning, limiting their utility in generating the global semantic features required for efficient image retrieval.

Self-supervised models display varied outcomes. 
DINOv2 shows retrieval capabilities comparable to supervised approaches.  
RAD-DINO, the medical adaptation of DINOv2, trained on chest X-ray images only, outperforms its natural domain counterpart.
Interestingly, this advantage does not extend consistently across all medical-specific variants, as MedSAM and MedCLIP do not surpass their natural image counterparts, SAM and CLIP, respectively.

Surprisingly, the MAE trained on natural images outperforms both DINO approaches, despite being trained only on natural images. 
Overall, it achieves the fourth-highest P@1 score.

In contrast to the foundation models, CVNet (Specialist) trained on the combined dataset, consistently outperforms all foundation models. CVNet-Global101 achieves a P@1 score of 0.650 (micro-averaged) and 0.266 (macro-averaged), significantly higher than any foundation model.

\begin{table*}[t!]
\centering
    \caption{Retrieval performance of foundation models on the combined dataset by modality. 
    }
    \label{tab:retrieval_by_mod}
\begin{tabular}{c|ccccc|ccccc}
\toprule
                    & \multicolumn{5}{c|}{\textbf{micro P@1}}                                      & \multicolumn{5}{c}{\textbf{macro P@1}}                                        \\ 
                     & \textbf{CT}    & \textbf{MR} & \textbf{US} & \textbf{XR} &  \textbf{\textit{Avg.}}     & \textbf{CT}    & \textbf{MR} & \textbf{US} & \textbf{XR} & \textbf{\textit{Avg.}}      \\
\midrule
\multicolumn{11}{c}{\textbf{Off-the-shelf}}          \\
\midrule
ResNet & {\cellcolor[HTML]{FCAA5F}} \color[HTML]{000000} 0.560 & {\cellcolor[HTML]{8ECF67}} \color[HTML]{000000} 0.482 & {\cellcolor[HTML]{EBF7A3}} \color[HTML]{000000} 0.714 & {\cellcolor[HTML]{AF0926}} \color[HTML]{F1F1F1} 0.370 & {\cellcolor[HTML]{FFF7B2}} \color[HTML]{000000} 0.539 & {\cellcolor[HTML]{F7844E}} \color[HTML]{F1F1F1} 0.210 & {\cellcolor[HTML]{BFE47A}} \color[HTML]{000000} 0.185 & {\cellcolor[HTML]{F2FAAE}} \color[HTML]{000000} 0.497 & {\cellcolor[HTML]{D22B27}} \color[HTML]{F1F1F1} 0.076 & {\cellcolor[HTML]{EEF8A8}} \color[HTML]{000000} 0.203 \\
ViT & {\cellcolor[HTML]{FFF1A8}} \color[HTML]{000000} 0.577 & {\cellcolor[HTML]{48AE5C}} \color[HTML]{F1F1F1} 0.495 & {\cellcolor[HTML]{6BBF64}} \color[HTML]{000000} 0.769 & {\cellcolor[HTML]{CA2427}} \color[HTML]{F1F1F1} 0.371 & {\cellcolor[HTML]{B3DF72}} \color[HTML]{000000} 0.560 & {\cellcolor[HTML]{FCA55D}} \color[HTML]{000000} 0.213 & {\cellcolor[HTML]{54B45F}} \color[HTML]{F1F1F1} 0.201 & {\cellcolor[HTML]{87CB67}} \color[HTML]{000000} 0.549 & {\cellcolor[HTML]{B71126}} \color[HTML]{F1F1F1} 0.072 & {\cellcolor[HTML]{A5D86A}} \color[HTML]{000000} 0.217 \\
Ark & {\cellcolor[HTML]{91D068}} \color[HTML]{000000} 0.606 & {\cellcolor[HTML]{07753E}} \color[HTML]{F1F1F1} 0.511 & {\cellcolor[HTML]{08773F}} \color[HTML]{F1F1F1} 0.810 & {\cellcolor[HTML]{006837}} \color[HTML]{F1F1F1} 0.395 & {\cellcolor[HTML]{128A49}} \color[HTML]{F1F1F1} 0.586 & {\cellcolor[HTML]{54B45F}} \color[HTML]{F1F1F1} 0.247 & {\cellcolor[HTML]{036E3A}} \color[HTML]{F1F1F1} 0.216 & {\cellcolor[HTML]{036E3A}} \color[HTML]{F1F1F1} 0.607 & {\cellcolor[HTML]{006837}} \color[HTML]{F1F1F1} 0.137 & {\cellcolor[HTML]{006837}} \color[HTML]{F1F1F1} 0.245 \\
SAM & {\cellcolor[HTML]{A50026}} \color[HTML]{F1F1F1} 0.529 & {\cellcolor[HTML]{DAF08D}} \color[HTML]{000000} 0.465 & {\cellcolor[HTML]{F2FAAE}} \color[HTML]{000000} 0.709 & {\cellcolor[HTML]{F67C4A}} \color[HTML]{F1F1F1} 0.375 & {\cellcolor[HTML]{FDAF62}} \color[HTML]{000000} 0.522 & {\cellcolor[HTML]{A50026}} \color[HTML]{F1F1F1} 0.195 & {\cellcolor[HTML]{EEF8A8}} \color[HTML]{000000} 0.176 & {\cellcolor[HTML]{FFFCBA}} \color[HTML]{000000} 0.486 & {\cellcolor[HTML]{C41E27}} \color[HTML]{F1F1F1} 0.074 & {\cellcolor[HTML]{FEEB9D}} \color[HTML]{000000} 0.193 \\
MedSAM & {\cellcolor[HTML]{FEDC88}} \color[HTML]{000000} 0.570 & {\cellcolor[HTML]{A50026}} \color[HTML]{F1F1F1} 0.392 & {\cellcolor[HTML]{A50026}} \color[HTML]{F1F1F1} 0.587 & {\cellcolor[HTML]{EE613E}} \color[HTML]{F1F1F1} 0.374 & {\cellcolor[HTML]{A50026}} \color[HTML]{F1F1F1} 0.491 & {\cellcolor[HTML]{B30D26}} \color[HTML]{F1F1F1} 0.197 & {\cellcolor[HTML]{A50026}} \color[HTML]{F1F1F1} 0.127 & {\cellcolor[HTML]{A50026}} \color[HTML]{F1F1F1} 0.368 & {\cellcolor[HTML]{DE402E}} \color[HTML]{F1F1F1} 0.078 & {\cellcolor[HTML]{A50026}} \color[HTML]{F1F1F1} 0.154 \\
\midrule
CLIP & {\cellcolor[HTML]{F2FAAE}} \color[HTML]{000000} 0.585 & {\cellcolor[HTML]{0B7D42}} \color[HTML]{F1F1F1} 0.509 & {\cellcolor[HTML]{36A657}} \color[HTML]{F1F1F1} 0.785 & {\cellcolor[HTML]{F67A49}} \color[HTML]{F1F1F1} 0.375 & {\cellcolor[HTML]{73C264}} \color[HTML]{000000} 0.571 & {\cellcolor[HTML]{FDC776}} \color[HTML]{000000} 0.217 & {\cellcolor[HTML]{249D53}} \color[HTML]{F1F1F1} 0.206 & {\cellcolor[HTML]{78C565}} \color[HTML]{000000} 0.554 & {\cellcolor[HTML]{A50026}} \color[HTML]{F1F1F1} 0.069 & {\cellcolor[HTML]{87CB67}} \color[HTML]{000000} 0.222 \\
MedCLIP & {\cellcolor[HTML]{C5E67E}} \color[HTML]{000000} 0.597 & {\cellcolor[HTML]{A9DA6C}} \color[HTML]{000000} 0.477 & {\cellcolor[HTML]{B9E176}} \color[HTML]{000000} 0.739 & {\cellcolor[HTML]{FCAA5F}} \color[HTML]{000000} 0.377 & {\cellcolor[HTML]{C5E67E}} \color[HTML]{000000} 0.557 & {\cellcolor[HTML]{F98E52}} \color[HTML]{F1F1F1} 0.211 & {\cellcolor[HTML]{CFEB85}} \color[HTML]{000000} 0.182 & {\cellcolor[HTML]{BBE278}} \color[HTML]{000000} 0.527 & {\cellcolor[HTML]{1B9950}} \color[HTML]{F1F1F1} 0.130 & {\cellcolor[HTML]{CFEB85}} \color[HTML]{000000} 0.210 \\
BiomedCLIP & {\cellcolor[HTML]{006837}} \color[HTML]{F1F1F1} 0.635 & {\cellcolor[HTML]{108647}} \color[HTML]{F1F1F1} 0.507 & {\cellcolor[HTML]{006837}} \color[HTML]{F1F1F1} 0.817 & {\cellcolor[HTML]{A50026}} \color[HTML]{F1F1F1} 0.369 & {\cellcolor[HTML]{006837}} \color[HTML]{F1F1F1} 0.594 & {\cellcolor[HTML]{006837}} \color[HTML]{F1F1F1} 0.259 & {\cellcolor[HTML]{06733D}} \color[HTML]{F1F1F1} 0.214 & {\cellcolor[HTML]{0F8446}} \color[HTML]{F1F1F1} 0.595 & {\cellcolor[HTML]{FDB768}} \color[HTML]{000000} 0.091 & {\cellcolor[HTML]{0C7F43}} \color[HTML]{F1F1F1} 0.240 \\
BMC-CLIP & {\cellcolor[HTML]{138C4A}} \color[HTML]{F1F1F1} 0.627 & {\cellcolor[HTML]{006837}} \color[HTML]{F1F1F1} 0.515 & {\cellcolor[HTML]{08773F}} \color[HTML]{F1F1F1} 0.809 & {\cellcolor[HTML]{AF0926}} \color[HTML]{F1F1F1} 0.370 & {\cellcolor[HTML]{04703B}} \color[HTML]{F1F1F1} 0.592 & {\cellcolor[HTML]{70C164}} \color[HTML]{000000} 0.245 & {\cellcolor[HTML]{138C4A}} \color[HTML]{F1F1F1} 0.210 & {\cellcolor[HTML]{148E4B}} \color[HTML]{F1F1F1} 0.591 & {\cellcolor[HTML]{E75337}} \color[HTML]{F1F1F1} 0.080 & {\cellcolor[HTML]{2DA155}} \color[HTML]{F1F1F1} 0.233 \\
\midrule
MAE & {\cellcolor[HTML]{FEE593}} \color[HTML]{000000} 0.573 & {\cellcolor[HTML]{016A38}} \color[HTML]{F1F1F1} 0.514 & {\cellcolor[HTML]{08773F}} \color[HTML]{F1F1F1} 0.809 & {\cellcolor[HTML]{EC5C3B}} \color[HTML]{F1F1F1} 0.374 & {\cellcolor[HTML]{6BBF64}} \color[HTML]{000000} 0.572 & {\cellcolor[HTML]{7AC665}} \color[HTML]{000000} 0.244 & {\cellcolor[HTML]{006837}} \color[HTML]{F1F1F1} 0.217 & {\cellcolor[HTML]{006837}} \color[HTML]{F1F1F1} 0.610 & {\cellcolor[HTML]{BD1726}} \color[HTML]{F1F1F1} 0.073 & {\cellcolor[HTML]{128A49}} \color[HTML]{F1F1F1} 0.238 \\
DINOv2 & {\cellcolor[HTML]{FEDE89}} \color[HTML]{000000} 0.571 & {\cellcolor[HTML]{69BE63}} \color[HTML]{F1F1F1} 0.489 & {\cellcolor[HTML]{A9DA6C}} \color[HTML]{000000} 0.746 & {\cellcolor[HTML]{DE402E}} \color[HTML]{F1F1F1} 0.373 & {\cellcolor[HTML]{DAF08D}} \color[HTML]{000000} 0.552 & {\cellcolor[HTML]{FEE08B}} \color[HTML]{000000} 0.221 & {\cellcolor[HTML]{4BB05C}} \color[HTML]{F1F1F1} 0.202 & {\cellcolor[HTML]{C3E67D}} \color[HTML]{000000} 0.523 & {\cellcolor[HTML]{B71126}} \color[HTML]{F1F1F1} 0.072 & {\cellcolor[HTML]{A5D86A}} \color[HTML]{000000} 0.217 \\
RAD-DINO & {\cellcolor[HTML]{B7E075}} \color[HTML]{000000} 0.600 & {\cellcolor[HTML]{89CC67}} \color[HTML]{000000} 0.483 & {\cellcolor[HTML]{84CA66}} \color[HTML]{000000} 0.760 & {\cellcolor[HTML]{FFF5AE}} \color[HTML]{000000} 0.381 & {\cellcolor[HTML]{9DD569}} \color[HTML]{000000} 0.564 & {\cellcolor[HTML]{B3DF72}} \color[HTML]{000000} 0.238 & {\cellcolor[HTML]{5DB961}} \color[HTML]{F1F1F1} 0.200 & {\cellcolor[HTML]{89CC67}} \color[HTML]{000000} 0.547 & {\cellcolor[HTML]{BFE47A}} \color[HTML]{000000} 0.113 & {\cellcolor[HTML]{6BBF64}} \color[HTML]{000000} 0.226 \\
\midrule
\multicolumn{11}{c}{\textbf{Specialist}}          \\
\midrule
CVNetGlobal50 & 0.671 & 0.545 & 0.925 & 0.408 & 0.644 & 0.240 & 0.221 & 0.743 & 0.080 & 0.251 \\
CVNetGlobal101 & 0.666 & 0.557 & 0.941 & 0.423 & 0.650 & 0.257 & 0.233 & 0.772 & 0.086 & 0.266 \\

\bottomrule
\end{tabular}
\end{table*}

\subsection{Modality-specific insights}
\label{sec:mod-spec}
Analyzing performance by modality (\cref{tab:retrieval_by_mod}) provides richer insights. General results conceal substantial modality-based variability: retrieval is strongest for US (P@1 up to 0.817), followed by CT (0.635), MR (0.507), and XR (0.395). The comparatively low XR performance likely stems from challenges inherent in interpreting two-dimensional projections, which can obscure critical anatomical details and subtle structural variations.
In contrast, MR and CT slices often exhibit clearer anatomic boundaries and intensity patterns, facilitating discriminative feature extraction.
Overall, BiomedCLIP and BMC-CLIP emerge as the top-performing foundation models across modalities, except for XR, where Ark is performing best. 
This enhanced performance for Ark, as well as RAD-DINO and MedCLIP, in XR images is presumably due to their exclusive pretraining on chest X-ray data. It is essential to note, however, that these XR-specific models had exposure to subsets of the XR evaluation data during their training, potentially inflating their measured retrieval performance on XR tasks. Detailed retrieval performance metrics on individual datasets are available in Appendix \cref{tab:retrieval_by_ds}.

Ark notably generalizes beyond XR, showing commendable performance in other imaging modalities, whereas RAD-DINO and MedCLIP display comparatively limited transferability. 
Notably, MAE, despite being trained only on natural images, performs on par with the best biomedical models on MR and US.

In line with the previously described superior performance of the specialist (CVNetGlobal50/101), also when separating the modalities, the specialist performs better than the foundation models.
However, also here we observe a much lower performance in XR than in the other modalities, likely due to the aforementioned difficulties for 2D projections.

\subsection{Pathological vs. anatomical structures}
\label{sec:imbalance}
Across all models, we notice a wide margin between class-wise (micro) and sample-wise (macro) scores indicating higher performance for more frequent classes on our combined dataset (\cref{tab:retrieval}, Appendix \cref{fig:p1vsSamplesPerClassInIndexBiomedCLIP}).
In this chapter, we investigate where this discrepancy comes from. 
To this end, we use BiomedCLIP exemplarily, since it is the best-performing foundation model and the phenomenon can be observed across the models. 

Our findings show superior performance for anatomical over pathological classes, highlighting the model's challenges in accurately retrieving pathological structures.
This is further supported by the quantitative analysis in Appendix \cref{tab:performance_retrieval_sep_pathology} where P@1 is at 0.812 for anatomical classes and only 0.451 on pathological classes.
Due to the imbalance between anatomical classes (24) and pathological classes (161), the macro average scores are dominated by the pathological classes.

\begin{table*}[t!]
    \centering
    \caption{
    Performance of the foundation models on the combined dataset using kNN and linear classification. 
    }
    \label{tab:classification}
    \begin{tabular}{c|cc|cc|cc|cc}
    \toprule
        \textbf{}           & \multicolumn{4}{c|}{\textbf{kNN}}   & \multicolumn{4}{c}{\textbf{Linear Probing}}                                                                                                                                                                                         \\ 
        \textbf{}           & \multicolumn{2}{c|}{\textbf{Micro}} & \multicolumn{2}{c|}{\textbf{Macro}}     & \multicolumn{2}{c|}{\textbf{Micro}} & \multicolumn{2}{c}{\textbf{Macro}}                                                                                                          \\ 
        \textbf{}           & \textbf{AURPC}                      & \multicolumn{1}{c|}{\textbf{F1}}        & \textbf{AURPC}                      & \textbf{F1}                        & \textbf{AURPC}     & \multicolumn{1}{c|}{\textbf{F1}}        & \textbf{AURPC}     & \textbf{F1}        \\ 

\midrule
\multicolumn{9}{c}{\textbf{Off-the-shelf}}          \\
\midrule
ResNet & {\cellcolor[HTML]{F5FBB2}} \color[HTML]{000000} 0.635 & {\cellcolor[HTML]{F2FAAE}} \color[HTML]{000000} 0.620 & {\cellcolor[HTML]{EFF8AA}} \color[HTML]{000000} 0.213 & {\cellcolor[HTML]{FFFCBA}} \color[HTML]{000000} 0.190 & {\cellcolor[HTML]{63BC62}} \color[HTML]{F1F1F1} 0.706 & {\cellcolor[HTML]{57B65F}} \color[HTML]{F1F1F1} 0.648 & {\cellcolor[HTML]{3FAA59}} \color[HTML]{F1F1F1} 0.278 & {\cellcolor[HTML]{128A49}} \color[HTML]{F1F1F1} 0.262 \\
ViT & {\cellcolor[HTML]{DDF191}} \color[HTML]{000000} 0.645 & {\cellcolor[HTML]{BBE278}} \color[HTML]{000000} 0.633 & {\cellcolor[HTML]{D7EE8A}} \color[HTML]{000000} 0.220 & {\cellcolor[HTML]{A9DA6C}} \color[HTML]{000000} 0.210 & {\cellcolor[HTML]{16914D}} \color[HTML]{F1F1F1} 0.734 & {\cellcolor[HTML]{138C4A}} \color[HTML]{F1F1F1} 0.668 & {\cellcolor[HTML]{0B7D42}} \color[HTML]{F1F1F1} 0.300 & {\cellcolor[HTML]{036E3A}} \color[HTML]{F1F1F1} 0.275 \\
Ark & {\cellcolor[HTML]{17934E}} \color[HTML]{F1F1F1} 0.697 & {\cellcolor[HTML]{18954F}} \color[HTML]{F1F1F1} 0.658 & {\cellcolor[HTML]{0B7D42}} \color[HTML]{F1F1F1} 0.264 & {\cellcolor[HTML]{006837}} \color[HTML]{F1F1F1} 0.239 & {\cellcolor[HTML]{006837}} \color[HTML]{F1F1F1} 0.757 & {\cellcolor[HTML]{006837}} \color[HTML]{F1F1F1} 0.683 & {\cellcolor[HTML]{006837}} \color[HTML]{F1F1F1} 0.310 & {\cellcolor[HTML]{0E8245}} \color[HTML]{F1F1F1} 0.266 \\
SAM & {\cellcolor[HTML]{FDAF62}} \color[HTML]{000000} 0.599 & {\cellcolor[HTML]{FDC372}} \color[HTML]{000000} 0.601 & {\cellcolor[HTML]{FEEA9B}} \color[HTML]{000000} 0.199 & {\cellcolor[HTML]{FED07E}} \color[HTML]{000000} 0.179 & {\cellcolor[HTML]{FBA35C}} \color[HTML]{000000} 0.571 & {\cellcolor[HTML]{FDBF6F}} \color[HTML]{000000} 0.556 & {\cellcolor[HTML]{FDAD60}} \color[HTML]{000000} 0.163 & {\cellcolor[HTML]{F88950}} \color[HTML]{F1F1F1} 0.113 \\
MedSAM & {\cellcolor[HTML]{A50026}} \color[HTML]{F1F1F1} 0.550 & {\cellcolor[HTML]{A50026}} \color[HTML]{F1F1F1} 0.567 & {\cellcolor[HTML]{A50026}} \color[HTML]{F1F1F1} 0.146 & {\cellcolor[HTML]{A50026}} \color[HTML]{F1F1F1} 0.143 & {\cellcolor[HTML]{A50026}} \color[HTML]{F1F1F1} 0.497 & {\cellcolor[HTML]{A50026}} \color[HTML]{F1F1F1} 0.492 & {\cellcolor[HTML]{A50026}} \color[HTML]{F1F1F1} 0.100 & {\cellcolor[HTML]{A50026}} \color[HTML]{F1F1F1} 0.060 \\
\midrule
CLIP & {\cellcolor[HTML]{7DC765}} \color[HTML]{000000} 0.673 & {\cellcolor[HTML]{5AB760}} \color[HTML]{F1F1F1} 0.649 & {\cellcolor[HTML]{70C164}} \color[HTML]{000000} 0.243 & {\cellcolor[HTML]{5DB961}} \color[HTML]{F1F1F1} 0.221 & {\cellcolor[HTML]{0D8044}} \color[HTML]{F1F1F1} 0.743 & {\cellcolor[HTML]{0B7D42}} \color[HTML]{F1F1F1} 0.674 & {\cellcolor[HTML]{026C39}} \color[HTML]{F1F1F1} 0.308 & {\cellcolor[HTML]{006837}} \color[HTML]{F1F1F1} 0.278 \\
MedCLIP & {\cellcolor[HTML]{FBFDBA}} \color[HTML]{000000} 0.633 & {\cellcolor[HTML]{ECF7A6}} \color[HTML]{000000} 0.622 & {\cellcolor[HTML]{FBFDBA}} \color[HTML]{000000} 0.209 & {\cellcolor[HTML]{C1E57B}} \color[HTML]{000000} 0.205 & {\cellcolor[HTML]{39A758}} \color[HTML]{F1F1F1} 0.720 & {\cellcolor[HTML]{279F53}} \color[HTML]{F1F1F1} 0.660 & {\cellcolor[HTML]{249D53}} \color[HTML]{F1F1F1} 0.286 & {\cellcolor[HTML]{3CA959}} \color[HTML]{F1F1F1} 0.246 \\
BiomedCLIP & {\cellcolor[HTML]{006837}} \color[HTML]{F1F1F1} 0.711 & {\cellcolor[HTML]{016A38}} \color[HTML]{F1F1F1} 0.667 & {\cellcolor[HTML]{006837}} \color[HTML]{F1F1F1} 0.269 & {\cellcolor[HTML]{16914D}} \color[HTML]{F1F1F1} 0.231 & {\cellcolor[HTML]{4EB15D}} \color[HTML]{F1F1F1} 0.713 & {\cellcolor[HTML]{4BB05C}} \color[HTML]{F1F1F1} 0.651 & {\cellcolor[HTML]{51B35E}} \color[HTML]{F1F1F1} 0.273 & {\cellcolor[HTML]{6EC064}} \color[HTML]{000000} 0.232 \\
BMC-CLIP & {\cellcolor[HTML]{05713C}} \color[HTML]{F1F1F1} 0.708 & {\cellcolor[HTML]{006837}} \color[HTML]{F1F1F1} 0.667 & {\cellcolor[HTML]{036E3A}} \color[HTML]{F1F1F1} 0.268 & {\cellcolor[HTML]{219C52}} \color[HTML]{F1F1F1} 0.229 & {\cellcolor[HTML]{07753E}} \color[HTML]{F1F1F1} 0.749 & {\cellcolor[HTML]{0A7B41}} \color[HTML]{F1F1F1} 0.675 & {\cellcolor[HTML]{0C7F43}} \color[HTML]{F1F1F1} 0.300 & {\cellcolor[HTML]{54B45F}} \color[HTML]{F1F1F1} 0.239 \\
\midrule
MAE & {\cellcolor[HTML]{6BBF64}} \color[HTML]{000000} 0.677 & {\cellcolor[HTML]{6BBF64}} \color[HTML]{000000} 0.646 & {\cellcolor[HTML]{1B9950}} \color[HTML]{F1F1F1} 0.257 & {\cellcolor[HTML]{4EB15D}} \color[HTML]{F1F1F1} 0.223 & {\cellcolor[HTML]{3CA959}} \color[HTML]{F1F1F1} 0.719 & {\cellcolor[HTML]{36A657}} \color[HTML]{F1F1F1} 0.656 & {\cellcolor[HTML]{7AC665}} \color[HTML]{000000} 0.261 & {\cellcolor[HTML]{BFE47A}} \color[HTML]{000000} 0.201 \\
DINOv2 & {\cellcolor[HTML]{D3EC87}} \color[HTML]{000000} 0.649 & {\cellcolor[HTML]{C9E881}} \color[HTML]{000000} 0.630 & {\cellcolor[HTML]{BBE278}} \color[HTML]{000000} 0.227 & {\cellcolor[HTML]{C9E881}} \color[HTML]{000000} 0.204 & {\cellcolor[HTML]{33A456}} \color[HTML]{F1F1F1} 0.722 & {\cellcolor[HTML]{2AA054}} \color[HTML]{F1F1F1} 0.659 & {\cellcolor[HTML]{279F53}} \color[HTML]{F1F1F1} 0.285 & {\cellcolor[HTML]{0B7D42}} \color[HTML]{F1F1F1} 0.268 \\
RAD-DINO & {\cellcolor[HTML]{CBE982}} \color[HTML]{000000} 0.651 & {\cellcolor[HTML]{A2D76A}} \color[HTML]{000000} 0.637 & {\cellcolor[HTML]{C5E67E}} \color[HTML]{000000} 0.225 & {\cellcolor[HTML]{4EB15D}} \color[HTML]{F1F1F1} 0.223 & {\cellcolor[HTML]{148E4B}} \color[HTML]{F1F1F1} 0.736 & {\cellcolor[HTML]{16914D}} \color[HTML]{F1F1F1} 0.666 & {\cellcolor[HTML]{39A758}} \color[HTML]{F1F1F1} 0.280 & {\cellcolor[HTML]{7DC765}} \color[HTML]{000000} 0.227 \\
\midrule
\multicolumn{9}{c}{\textbf{Specialist}}          \\
\midrule
CVNetGlobal50 & 0.743 & 0.703 & 0.273 & 0.246 & 0.772 & 0.699 & 0.307 & 0.277 \\
CVNetGlobal101 & 0.735 & 0.703 & 0.283 & 0.256 & 0.767 & 0.702 & 0.319 & 0.296 \\

\bottomrule
    \end{tabular}
\end{table*}

\begin{figure}[!h]    
    \includegraphics[width=.45\textwidth]{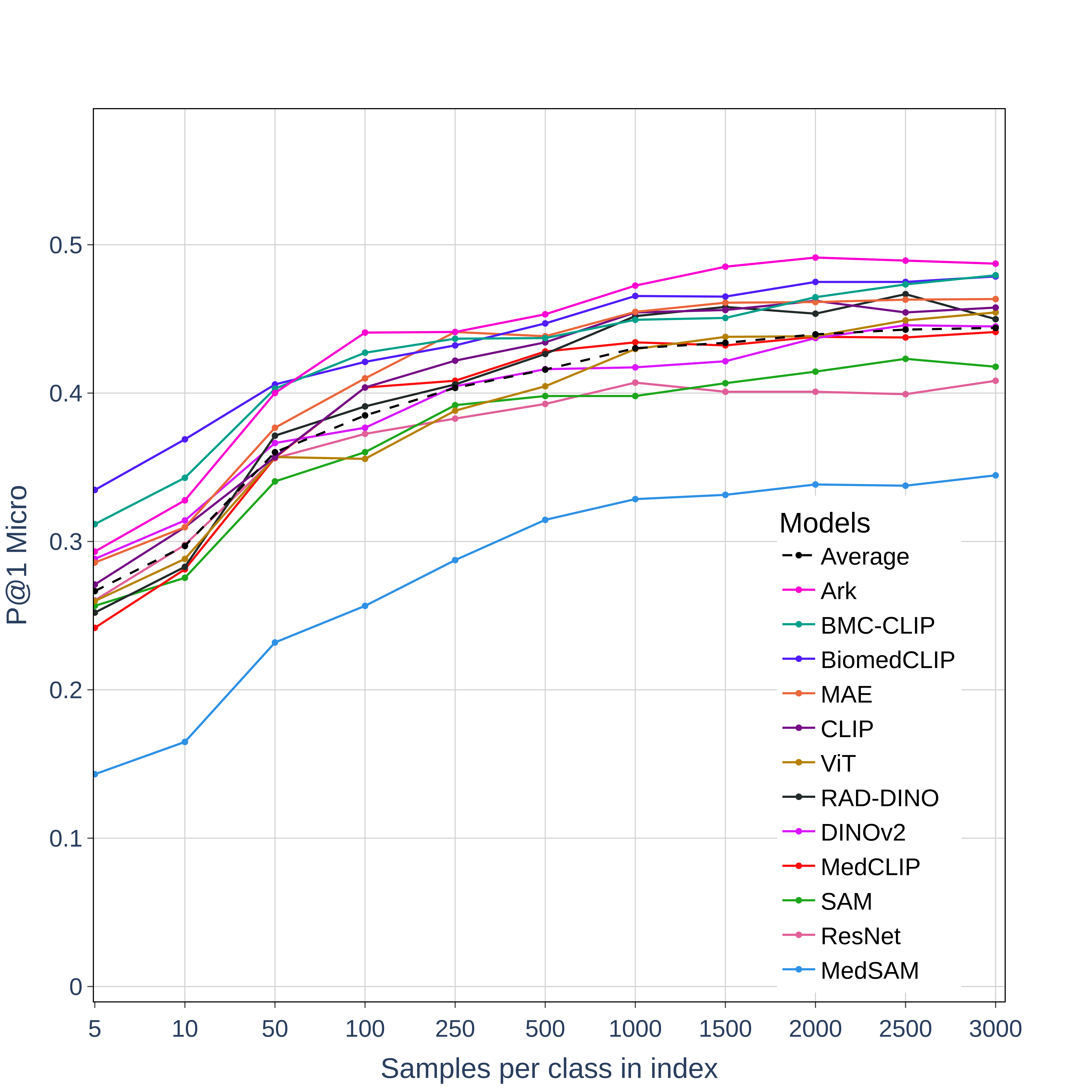}
    \caption{Effect of the number of samples per class in the index on the retrieval performance.
    Sample-wise performance ($P@1$ micro) of the foundation models with varying numbers of samples per class in the index. Performance saturates with 1000 samples.
    }
    \label{fig:pAt1vsIndex}
\end{figure}

\subsection{Effect of index size on retrieval performance}
Furthermore, in Appendix \cref{fig:p1vsSamplesPerClassInIndexBiomedCLIP} we show that classes with more samples in the index yield a higher P@1.
This discrepancy likely arises from a broader variety of images, covering more variations of the same class. 
This increases the chance of having images in the index that are very similar to the query image. 
To investigate whether retrieval performance reaches saturation with a specific number of samples per class in the index, we conduct an ablation study focusing on classes with over 3000 samples, comprising 64 classes (42 pathological and 22 anatomical). 
We standardize the query dataset size to the smallest class, containing 38 samples per class. 
Then, we randomly select $N$ cases per class, where $N$ ranges from 5 to 3000, to populate the index and conduct retrievals using the fixed query set.
For the best models, P@1 starts saturating at around 1000 samples per class in the index (\cref{fig:pAt1vsIndex}).
This finding suggests that up to 1000 samples per class, providing more samples improves the retrieval performance. 
However, beyond 1000 samples, additional samples do not significantly improve retrieval outcomes, pointing to a need for more advanced models for performance gains.

\subsection{Analysis of the embedding space}

In content-based image retrieval (CBIR), global features are essential for both the initial retrieval process and subsequent reranking, where similar images are refined based on finer-grained features. 
The quality of these global features determines how effectively similar images can be clustered, making the embedding space a crucial factor in retrieval performance \cite{shao2023global}. 
To gain deeper insights into the quality of these global features, we employ k-nearest neighbors (kNN) classification to assess clustering and linear probing (see \cref{sec:training_details_linear} for training details) to evaluate linear separability of class-relevant features. These analyses provide insights into how well the models capture meaningful representations for medical image retrieval.
The results are shown in \cref{tab:classification}.

BiomedCLIP demonstrates the best kNN classification performance (AURPC 0.711, F1 0.667) closely followed by BMC-CLIP, indicating effective clustering of medical images, including pathological structures. This suggests that BiomedCLIP’s and BMC-CLIP's global features capture relevant medical semantics. CLIP and MAE also perform well, showing their generalization ability to medical images despite being trained on natural images. In contrast, SAM and MedSAM show weaker clustering performance (AURPC 0.588 and 0.561), likely due to their focus on fine-grained segmentation, which does not translate well to global feature clustering.
These results are well aligned with the retrieval performance (\cref{tab:retrieval}).

For linear probing, Ark achieves the highest performance (AURPC 0.757, F1 0.683), showing superior linear separability of features. 
BMC-CLIP achieves the second highest linear probing scores. 
While BiomedCLIP is superior in clustering, its slightly lower linear probing scores (AURPC 0.713) suggest that its feature space is semantically clustered but class-relevant features are less preserved than for Ark and BMC-CLIP. 
CLIP achieves the third highest linear probing score, despite being trained on natural images only, outperforming even some medical models.
All self-supervised models show good linear probing results, indicating they capture rich, discriminative features.

\begin{figure*}[h]
    \centering
    \includegraphics[width=.8\textwidth]{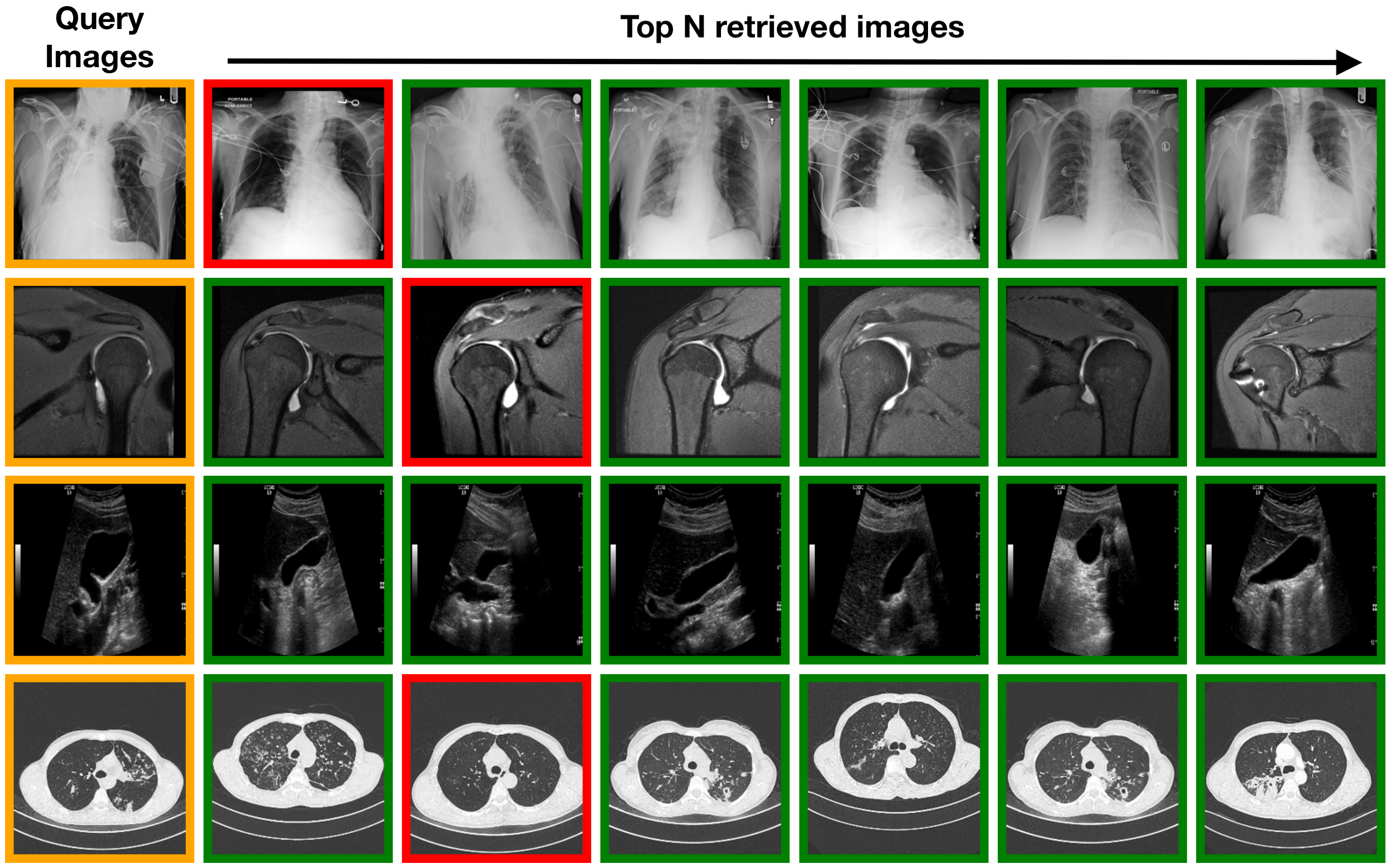}
    \caption{
    Qualitative results of image retrieval using BiomedCLIP and feature extractor.
    Rows display the query image highlighted in orange on the left, followed by the six most similar images, sorted by descending similarity. Green frames indicate a match with the query image's class, red indicates a mismatch.
    }
    \label{fig:qualitative}
\end{figure*}

\begin{figure*}[h]
    \includegraphics[width=\textwidth]{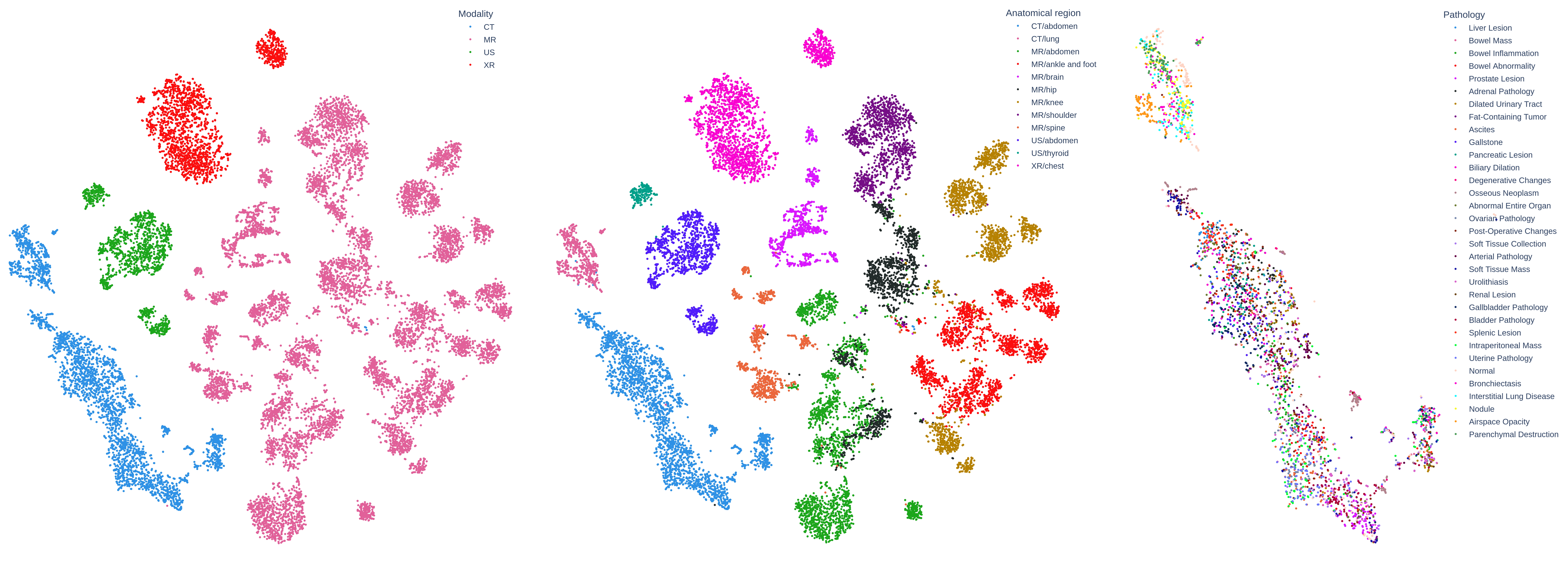}
    \caption{
    t-SNE projections of the BiomedCLIP embeddings of the combined dataset. 
    Left: Modalities are color-coded
    Center: Anatomical structures are color-coded.
    Right: Filtered for CT images only. Classes are color-coded.
}\label{fig:tsne}
\end{figure*}

\subsection{Qualitative results}
To complement the quantitative evaluations, we present qualitative results that highlight the retrieval performance and embedding space organization of BiomedCLIP. These visualizations provide deeper insights into how the model performs across different modalities and classes.

\subsubsection{Retrieval Examples}
\cref{fig:qualitative} showcases examples of query images alongside their top retrieved images using BiomedCLIP. 
The retrieved results are color-coded, with green frames indicating correct matches (same class as the query) and red frames marking incorrect matches. 
The examples span across all our evaluated imaging modalities (XR, MRI, US, CT). Notably, the model demonstrates high retrieval accuracy for most queries. 
While there are high variations in rotation, mirroring and translation, the model is capable of extracting the class-relevant features and, in most cases, correctly retrieving the respective samples from the index. 
These results emphasize the model's capability to generalize well across modalities.

\subsubsection{Embedding Space Visualization}
In \cref{fig:tsne}, we provide a t-SNE\cite{t-SNE} projection of the BiomedCLIP embeddings for the combined dataset, visualized with three color encodings: (1) imaging modality, (2) anatomical region, and (3) class labels. 
The embeddings reveal well-separated clusters for modality and anatomy regions, indicating that the model effectively learned representations that capture modality-specific and anatomical information. Distinct clustering of modalities (CT, MRI, XR, US) suggests that the model can discern key features unique to each modality. Furthermore, within modalities, embeddings organized by anatomical region show clear separability, reflecting the model's understanding of structural and contextual features. However, when zooming into the CT cluster, pathological clusters are less well separated. The dense overlapping of some pathology classes highlights the ongoing challenge of distinguishing subtle pathological features in visually similar images.

\section{Discussion}
This study explores the potential of vision foundation models as feature extractors for content-based medical image retrieval (CBIR) in radiology, providing a detailed benchmark across diverse modalities, models, and retrieval scenarios. 
Our findings underscore both the promise and limitations of foundation models in addressing the challenges of general-purpose CBIR systems in radiology.

Compared to previous research, earlier studies using foundation models for CBIR \cite{3dMIR, bayerMIR} were limited in scope. They often focused on single modalities, specific organs, or excluded pathologies, failing to reflect a realistic and comprehensive radiology setting. Additionally, these studies lacked detailed evaluations, such as comparisons with trained specialists, the influence of index size, and analyses of the embedding space. In contrast, our work offers a thorough evaluation, providing valuable insights into the retrieval capabilities of foundation models in radiology.

We show that weakly-supervised biomedical foundation models, particularly BiomedCLIP and BMC-CLIP, demonstrate strong generalization capabilities, achieving competitive retrieval performance (micro P@1 of 0.594) without task-specific fine-tuning (\cref{sec:retrieval}).  
In contrast, our trained specialist (CVNetGlobal101) achieves superior performance (micro P@1 of 0.650) through dedicated training.
While task-specific models (specialists) achieve higher retrieval performance, they require extensive labeled data and computational resources. Foundation models, used off-the-shelf, present a scalable alternative, particularly in data-sparse scenarios or resource-constrained environments.

Retrieval performance varied significantly across modalities, with ultrasound (US) achieving the highest accuracy (micro P@1 of 0.817) and X-ray (XR) the lowest (micro P@1 of 0.395), reflecting the challenges of 2D projection images. While BiomedCLIP is superior to other foundation models across CT, MR, and US modalities, Ark outperformed in XR.

Furthermore, we show that pathological features were notably harder to retrieve compared to anatomical features (P@1 of 0.451 vs. 0.812). This is likely due to the aforementioned subtle and nuanced structures of pathologies, which can easily be overshadowed by the anatomical similarities that are more pronounced and consistent across different images. 
Anatomical features tend to have more defined and recognizable patterns that are easier for models to capture and match in a retrieval context. 
In contrast, pathological features often vary greatly in appearance and can be subtle or less distinct, making them harder to accurately detect and retrieve solely based on visual similarity. 
This is further supported by our qualitative analysis using t-SNE (\cref{fig:tsne}), where anatomies can be clearly distinguished, whereas pathologies can not.

Our linear probing experiments revealed that Ark is superior to all other foundation models, indicating that it preserves class-relevant features the most. Future work should investigate how to leverage those superior global features. This could be done by employing re-ranking methods such as in \cite{cvnet}, where the global features are used for local refinement.

\subsection{Limitations}
Despite its contributions, our study has limitations. 
Our combined dataset is limited to single-class labels only, which fails to capture the co-occurrence of multiple pathologies, a common scenario in clinical practice. 
This limitation impacts the realism of the benchmark and its applicability to real-world scenarios. 
Furthermore, the evaluation metrics rely on class labels, which can oversimplify the concept of visual similarity. 
Pathologies with distinct visual appearances may share the same label, introducing ambiguity into the retrieval evaluation.
Additionally, by treating slice-level images from CT and MR scans independently, our approach ignores crucial contextual information about the spatial continuity and locality of anatomical structures inherently present in the original 3D volume data, potentially reducing the effectiveness of the learned representations.

\subsection{Future work}
Future research should address these limitations while exploring new directions to advance CBIR systems. 
First, our results indicate that large-scale domain-specific text-supervised pretraining is superior to other training schemes and training data. 
Therefore, future foundation models that can be used in CBIR should be trained using the CLIP training paradigm, preferably with high-resolution and high-quality data relevant to radiology. 
Regarding the actual usage of foundation models for CBIR we see the following future work. 
When the deployment domain is clear and the training data exists, investigating fine-tuning foundation models for specific use cases, has the potential to enhance retrieval accuracy by tailoring models to specific clinical requirements.
Orthogonal to that, future work should investigate how foundation models' global representation can be used in the re-ranking stage of two-stage retrieval systems \cite{cvnet,DeepImageRetrieval}.
Alternatively, for unknown domains and tasks, future work should investigate how to alleviate the inferior performance in pathology retrieval. 
Similar to \cite{denner2024visual, redcircle, maskinversion, alphaclip, rezaei2024learning}, future work should investigate how to guide the model's attention to a region of interest (e.g. pathology), allowing better capturing of pathological structures in the embedding space, thereby increasing retrieval performance. 
Furthermore, this could help overcome the burden that off-the-shelf foundation models currently lack the ability to dynamically adapt their feature representation according to a similarity definition.

By addressing these limitations and pursuing these research directions, future work could significantly advance the utility and applicability of foundation models for CBIR in radiology, making them more versatile and effective in diverse clinical and research contexts.

\section{Conclusion}
This study demonstrates the promise of vision foundation models as general-purpose feature extractors for content-based medical image retrieval in radiology. BiomedCLIP showed competitive performance across multiple modalities without task-specific fine-tuning, positioning foundation models as scalable and practical alternatives in data-constrained settings. While specialized CBIR systems still outperform in terms of accuracy, the versatility and ease of deployment of foundation models pave the way for advancing scalable, generalizable retrieval systems in radiology.

\printcredits

\section*{Declaration of Generative AI and AI-assisted technologies in the writing process}
During the preparation of this work the author(s) used ChatGPT in
order to improve readability and language. After using this tool/service, the author(s) reviewed and edited the content as needed and take(s) full responsibility for the content of the publication.

\section*{Declaration of competing interest}
The authors declare that they have no known competing financial
interests or personal relationships that could have appeared to influence the work reported in this paper.

\section*{Acknowledgment}
This work was supported by the NUM 2.0 (FKZ: 01KX2121).

\bibliography{bib}

\clearpage

\appendix

\section{CVNet training details}
\label{sec:training_details_cvnet}
We follow the implementation from CVNet \cite{cvnet}.
We use ResNet-50 and ResNet-101 as the encoder with  ImageNet\cite{Imagenet} pretrained weights.
Global descriptor size is 2048. We set $\tau$ to 1/30, $m$ to 0.15, $\eta$  to 0.999, and $\lambda_{cls}$ and $\lambda_{con}$ to 0.5. The models are trained for 25 epochs on the training dataset,
using a learning rate of 0.005, and a batch size of 224 and 128 for ResNet-50 and ResNet-101, respectively.
We base our implementation on \href{https://github.com/edwardguil/SuperCVNet}{https://github.com/edwardguil/SuperCVNet}.

\section{Linear classifier training details}
\label{sec:training_details_linear}
The training details for the linear classifier are outlined as follows: The model was a single linear layer without an activation function. It was trained for a total of 100 epochs, with an early stopping mechanism in place after 20 epochs if no further improvements were observed. A learning rate of 0.001  was employed throughout the training process. The optimizer utilized was AdamW\cite{loshchilov2017fixing}.

\section{Comparison to a specialized medical image retrieval system.}
\label{sec:contextualize_performance}

To contextualize the efficacy of employing vision foundation models as feature extractors, it is critical to benchmark their performance against image retrieval systems designed for a narrower spectrum of pathologies. 
We identified only one publicly available radiological image retrieval system that operates without requiring additional user input: X-MIR by Hu et al. \cite{XMIR}, which utilizes a DenseNet121 \cite{DenseNet}  trained on a COVID-19 chest X-ray dataset \cite{Covid-net} for the retrieval of pneumonia, Covid-19, and normal cases.

Our evaluation unfolds in two folds.
Firstly, we evaluate X-MIR on our combined dataset to highlight its limited generalizability to unseen classes (Appendix \cref{tab:retrieval_xmir}).
Secondly, we evaluate the best-performing foundation model on our combined dataset, BiomedCLIP,  on the COVID-19 chest X-ray dataset \cite{Covid-net}, which X-MIR was specifically trained on. Given updates to the dataset since its initial publication, we evaluate the X-MIR model on a subset of the original dataset, noting discrepancies from the originally reported results.
Despite BiomedCLIP not being fine-tuned on this dataset, it closely rivals X-MIR's specialized performance, underscoring the versatility and robustness of foundation models in broader medical image retrieval applications (Appendix \cref{tab:retrieval_covid}).

\begin{table}[!h]
\centering
\caption{Retrieval performance on the COVID-19 chest X-ray dataset.}
\label{tab:retrieval_covid}
\begin{tabular}{c|c|c|c|c}
\toprule
\textbf{Model} & \textbf{P@1} & \textbf{P@3} & \textbf{P@5} & \textbf{P@10} \\
\midrule
\multicolumn{5}{c}{\textbf{Sample-wise (micro)}} \\
\midrule
DenseNet      & 0.811 & 0.828 & 0.831 & 0.820 \\
BiomedCLIP    & 0.913 & 0.923 & 0.910 & 0.921 \\
X-MIR         & 0.954 & 0.934 & 0.934 & 0.932 \\
\midrule
\multicolumn{5}{c}{\textbf{Class-wise (macro)}} \\
\midrule
DenseNet      & 0.811 & 0.828 & 0.830 & 0.820 \\
BiomedCLIP    & 0.914 & 0.924 & 0.910 & 0.921 \\
X-MIR         & 0.954 & 0.933 & 0.934 & 0.931 \\
\bottomrule
\end{tabular}
\end{table}

\begin{table}[!h]
\centering
\caption{Retrieval performance of a specialist on the combined dataset}
\label{tab:retrieval_xmir}
\begin{tabular}{c|c|c|c|c}
\toprule
\textbf{Model} & \textbf{P@1} & \textbf{P@3} & \textbf{P@5} & \textbf{P@10} \\
\midrule
\multicolumn{5}{c}{\textbf{Sample-wise (micro)}} \\
\midrule
BiomedCLIP     & 0.594        & 0.590        & 0.588        & 0.583         \\
X-MIR          & 0.478        & 0.471        & 0.466        & 0.457         \\
\midrule
\multicolumn{5}{c}{\textbf{Class-wise (macro)}} \\
\midrule
BiomedCLIP     & 0.240        & 0.230        & 0.224        & 0.217         \\
X-MIR          & 0.149        & 0.142        & 0.138        & 0.131         \\
\bottomrule
\end{tabular}
\end{table}

\begin{figure}[!t]
   \includegraphics[width=.45\textwidth]{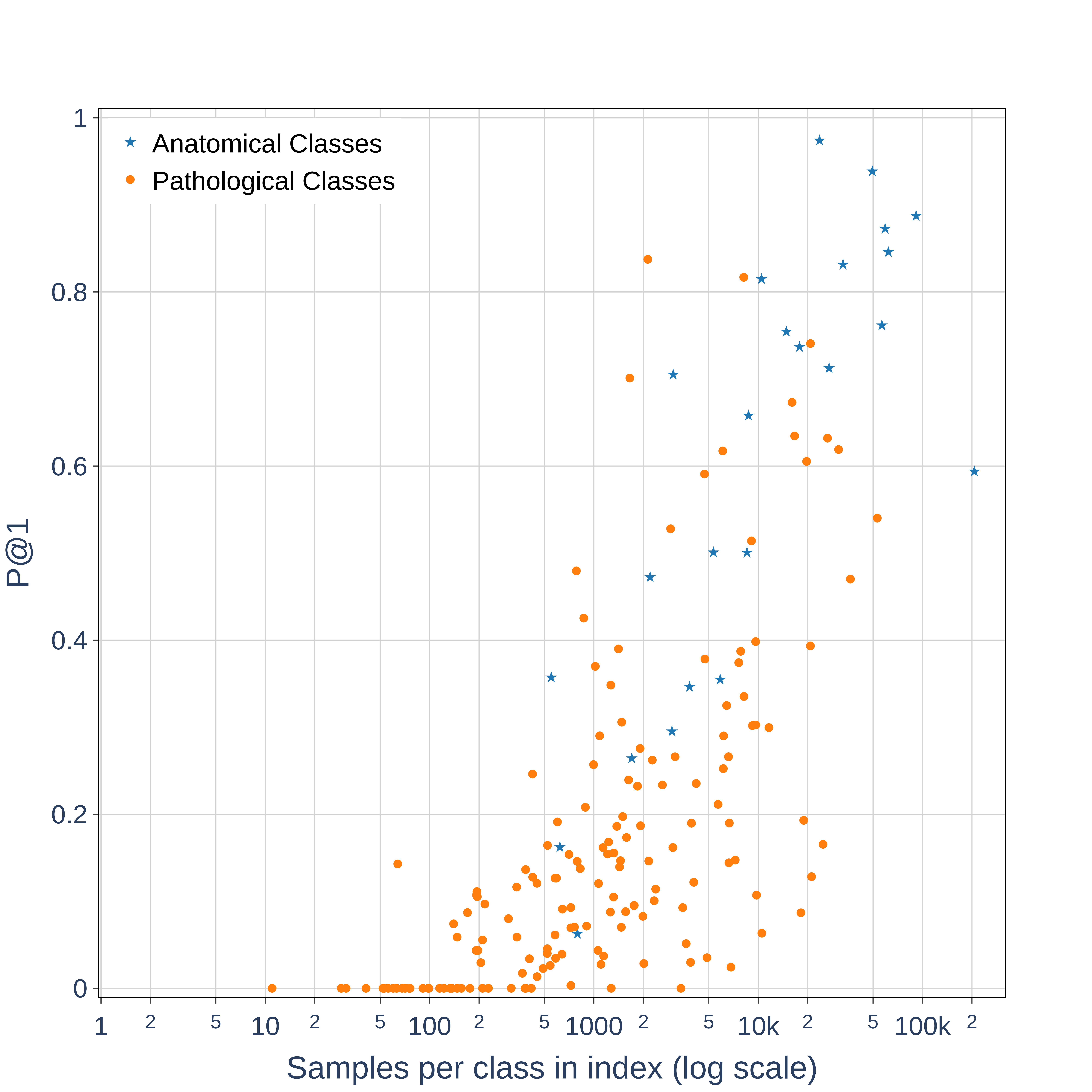}
    \caption{Discrepancy of retrieval performance between anatomical(blue) and pathological(orange) classes for BiomedCLIP considering the number of samples per class in the index. 
    }
\label{fig:p1vsSamplesPerClassInIndexBiomedCLIP}
\end{figure}

\begin{table*}
    \centering
    \caption{
        Datasets utilized for the benchmark.
        For image retrieval, the training dataset was used for indexing and the test dataset for querying/retrieval.
        For kNN classification, the validation dataset was used to determine the best $K$.
        For linear classification, the validation dataset was used for early stopping. 
    }
    \label{tab:datasets}
    \begin{tabular}{c|c|ccc|c}
        \textbf{Dataset}               & \textbf{Modality} & \textbf{Train} & \textbf{Val} & \textbf{Test} & \makecell{\textbf{Classes} \\ \#Path. + \#Anat.} \\ 
        \midrule
        NIH14 \cite{NIH14}             & XR                & 66,192         & 7,279        & 17,853        & 14+1                              \\ 
        MIMIC \cite{MIMIC}             & XR                & 229,597        & 1,775        & 2,473         & 12+1                              \\ 
        CheXpert \cite{CheXpert}       & XR                & 50,194         & 5,587        & 54            & 12+1                              \\ 
        RadImageNet \cite{RadImageNet} & MR                & 460,479        & 48,820       & 57,103        & 109+7                             \\
        \multicolumn{1}{l|}{}          & CT                & 202,521        & 24,660       & 63,198        & 32+2                              \\
        \multicolumn{1}{l|}{}          & US                & 313,569        & 27,417       & 31,072        & 14+1                              \\ 
        \midrule
        Total                          &                   & 1,322,552      & 115,538      & 171,753       & 161+24
    \end{tabular}
\end{table*}

\begin{figure*}
    \centering
    \includegraphics[width=.9\textwidth]{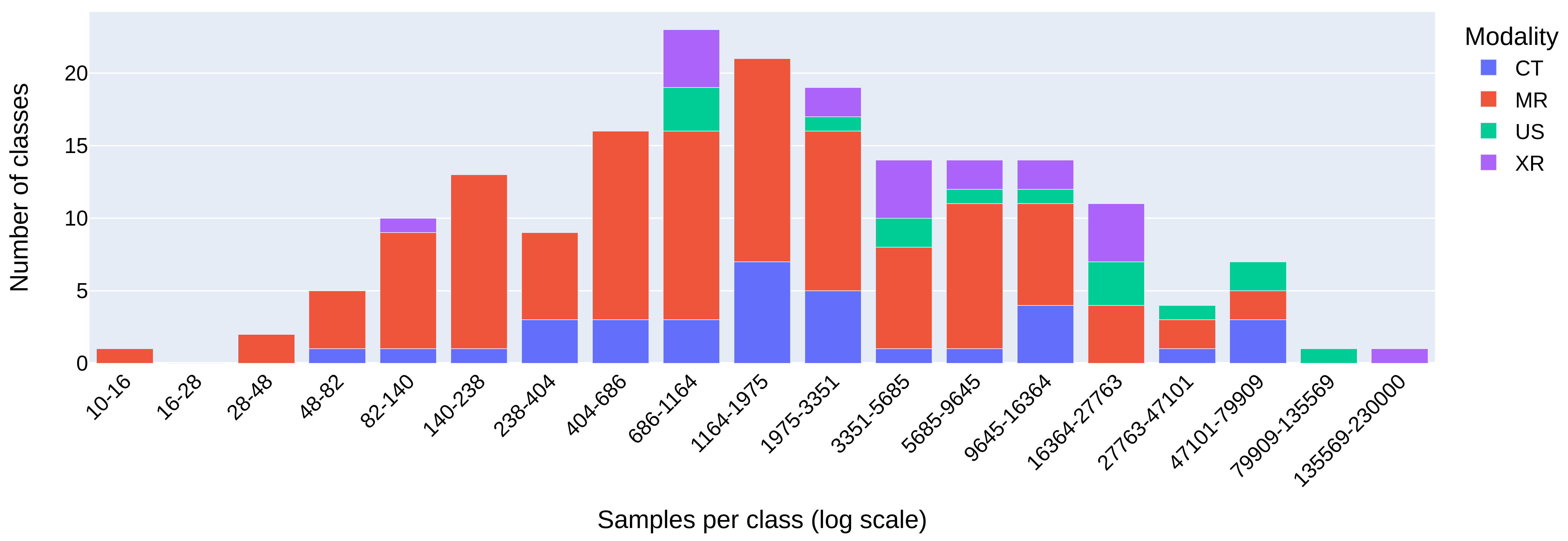}
    \caption{Class imbalance of the combined dataset. The x-axis shows the samples per class and the y-axis the number of classes.}
    \label{fig:imbalance}
\end{figure*}

\begin{table*}
\centering
\caption{Retrieval performance (micro P@1) of the foundation models on the combined datasets hierarchically separated by modality and then anatomical classes and pathological classes.}
\label{tab:performance_retrieval_sep_pathology}
\begin{tabular}{c|cc|cc|cc|cc|ccc} 
\toprule
Modality       & \multicolumn{2}{c|}{CT} & \multicolumn{2}{c|}{MR}  & \multicolumn{2}{c|}{US} & \multicolumn{2}{c|}{XR} & \multicolumn{3}{c}{AVG.} \\
Type & Anat. & Path. & Anat. & Path. & Anat. & Path. & Anat. & Path. & Anat. & Path. & Avg. \\
Classes & 2 & 32 & 7 & 109 & 14 & 1 & 1 & 14 & 24 & 161 & 185 \\
Samples & 33067 & 30131 & 14257 & 42846 & 28561 & 2511 & 10883 & 9497 & 86768 & 84985 & 171753 \\ 
\midrule
\multicolumn{12}{c}{\textbf{Off-the-shelf}} \\
\midrule                                             
ResNet & {\cellcolor[HTML]{FEE999}} \color[HTML]{000000} 0.595 & {\cellcolor[HTML]{FDB163}} \color[HTML]{000000} 0.522 & {\cellcolor[HTML]{93D168}} \color[HTML]{000000} 0.730 & {\cellcolor[HTML]{B9E176}} \color[HTML]{000000} 0.400 & {\cellcolor[HTML]{E8F59F}} \color[HTML]{000000} 0.730 & {\cellcolor[HTML]{E95538}} \color[HTML]{F1F1F1} 0.530 & {\cellcolor[HTML]{0B7D42}} \color[HTML]{F1F1F1} 0.635 & {\cellcolor[HTML]{A50026}} \color[HTML]{F1F1F1} 0.067 & {\cellcolor[HTML]{FFF7B2}} \color[HTML]{000000} 0.666 & {\cellcolor[HTML]{FED884}} \color[HTML]{000000} 0.410 & {\cellcolor[HTML]{FFF5AE}} \color[HTML]{000000} 0.539 \\
ViT & {\cellcolor[HTML]{ADDC6F}} \color[HTML]{000000} 0.647 & {\cellcolor[HTML]{A50026}} \color[HTML]{F1F1F1} 0.500 & {\cellcolor[HTML]{63BC62}} \color[HTML]{F1F1F1} 0.746 & {\cellcolor[HTML]{78C565}} \color[HTML]{000000} 0.411 & {\cellcolor[HTML]{6EC064}} \color[HTML]{000000} 0.785 & {\cellcolor[HTML]{FED481}} \color[HTML]{000000} 0.586 & {\cellcolor[HTML]{45AD5B}} \color[HTML]{F1F1F1} 0.625 & {\cellcolor[HTML]{DB382B}} \color[HTML]{F1F1F1} 0.080 & {\cellcolor[HTML]{91D068}} \color[HTML]{000000} 0.706 & {\cellcolor[HTML]{FEE08B}} \color[HTML]{000000} 0.411 & {\cellcolor[HTML]{B5DF74}} \color[HTML]{000000} 0.560 \\
Ark & {\cellcolor[HTML]{39A758}} \color[HTML]{F1F1F1} 0.682 & {\cellcolor[HTML]{FDB768}} \color[HTML]{000000} 0.523 & {\cellcolor[HTML]{6EC064}} \color[HTML]{000000} 0.743 & {\cellcolor[HTML]{006837}} \color[HTML]{F1F1F1} 0.434 & {\cellcolor[HTML]{06733D}} \color[HTML]{F1F1F1} 0.830 & {\cellcolor[HTML]{FECC7B}} \color[HTML]{000000} 0.582 & {\cellcolor[HTML]{FED07E}} \color[HTML]{000000} 0.583 & {\cellcolor[HTML]{006837}} \color[HTML]{F1F1F1} 0.179 & {\cellcolor[HTML]{2DA155}} \color[HTML]{F1F1F1} 0.728 & {\cellcolor[HTML]{199750}} \color[HTML]{F1F1F1} 0.441 & {\cellcolor[HTML]{138C4A}} \color[HTML]{F1F1F1} 0.586 \\
SAM & {\cellcolor[HTML]{A50026}} \color[HTML]{F1F1F1} 0.508 & {\cellcolor[HTML]{98D368}} \color[HTML]{000000} 0.552 & {\cellcolor[HTML]{E5F49B}} \color[HTML]{000000} 0.695 & {\cellcolor[HTML]{EBF7A3}} \color[HTML]{000000} 0.389 & {\cellcolor[HTML]{FFFCBA}} \color[HTML]{000000} 0.713 & {\cellcolor[HTML]{A2D76A}} \color[HTML]{000000} 0.669 & {\cellcolor[HTML]{026C39}} \color[HTML]{F1F1F1} 0.638 & {\cellcolor[HTML]{BE1827}} \color[HTML]{F1F1F1} 0.073 & {\cellcolor[HTML]{F36B42}} \color[HTML]{F1F1F1} 0.623 & {\cellcolor[HTML]{ECF7A6}} \color[HTML]{000000} 0.420 & {\cellcolor[HTML]{FDAF62}} \color[HTML]{000000} 0.522 \\
MedSAM & {\cellcolor[HTML]{FEDA86}} \color[HTML]{000000} 0.587 & {\cellcolor[HTML]{98D368}} \color[HTML]{000000} 0.552 & {\cellcolor[HTML]{A50026}} \color[HTML]{F1F1F1} 0.569 & {\cellcolor[HTML]{A50026}} \color[HTML]{F1F1F1} 0.333 & {\cellcolor[HTML]{A50026}} \color[HTML]{F1F1F1} 0.595 & {\cellcolor[HTML]{A50026}} \color[HTML]{F1F1F1} 0.489 & {\cellcolor[HTML]{199750}} \color[HTML]{F1F1F1} 0.630 & {\cellcolor[HTML]{DB382B}} \color[HTML]{F1F1F1} 0.080 & {\cellcolor[HTML]{A50026}} \color[HTML]{F1F1F1} 0.592 & {\cellcolor[HTML]{A50026}} \color[HTML]{F1F1F1} 0.387 & {\cellcolor[HTML]{A50026}} \color[HTML]{F1F1F1} 0.491 \\
\midrule
CLIP & {\cellcolor[HTML]{F2FAAE}} \color[HTML]{000000} 0.616 & {\cellcolor[HTML]{98D368}} \color[HTML]{000000} 0.552 & {\cellcolor[HTML]{006837}} \color[HTML]{F1F1F1} 0.790 & {\cellcolor[HTML]{5DB961}} \color[HTML]{F1F1F1} 0.415 & {\cellcolor[HTML]{63BC62}} \color[HTML]{F1F1F1} 0.788 & {\cellcolor[HTML]{006837}} \color[HTML]{F1F1F1} 0.745 & {\cellcolor[HTML]{0B7D42}} \color[HTML]{F1F1F1} 0.635 & {\cellcolor[HTML]{D02927}} \color[HTML]{F1F1F1} 0.077 & {\cellcolor[HTML]{9BD469}} \color[HTML]{000000} 0.704 & {\cellcolor[HTML]{57B65F}} \color[HTML]{F1F1F1} 0.436 & {\cellcolor[HTML]{75C465}} \color[HTML]{000000} 0.571 \\
MedCLIP & {\cellcolor[HTML]{5AB760}} \color[HTML]{F1F1F1} 0.673 & {\cellcolor[HTML]{EA5739}} \color[HTML]{F1F1F1} 0.512 & {\cellcolor[HTML]{E9F6A1}} \color[HTML]{000000} 0.692 & {\cellcolor[HTML]{96D268}} \color[HTML]{000000} 0.406 & {\cellcolor[HTML]{ABDB6D}} \color[HTML]{000000} 0.761 & {\cellcolor[HTML]{B30D26}} \color[HTML]{F1F1F1} 0.496 & {\cellcolor[HTML]{A50026}} \color[HTML]{F1F1F1} 0.550 & {\cellcolor[HTML]{026C39}} \color[HTML]{F1F1F1} 0.178 & {\cellcolor[HTML]{CBE982}} \color[HTML]{000000} 0.690 & {\cellcolor[HTML]{E5F49B}} \color[HTML]{000000} 0.421 & {\cellcolor[HTML]{C3E67D}} \color[HTML]{000000} 0.557 \\
BiomedCLIP & {\cellcolor[HTML]{006837}} \color[HTML]{F1F1F1} 0.711 & {\cellcolor[HTML]{98D368}} \color[HTML]{000000} 0.552 & {\cellcolor[HTML]{138C4A}} \color[HTML]{F1F1F1} 0.773 & {\cellcolor[HTML]{45AD5B}} \color[HTML]{F1F1F1} 0.418 & {\cellcolor[HTML]{006837}} \color[HTML]{F1F1F1} 0.836 & {\cellcolor[HTML]{FEEC9F}} \color[HTML]{000000} 0.602 & {\cellcolor[HTML]{FDFEBC}} \color[HTML]{000000} 0.595 & {\cellcolor[HTML]{FEDC88}} \color[HTML]{000000} 0.111 & {\cellcolor[HTML]{006837}} \color[HTML]{F1F1F1} 0.748 & {\cellcolor[HTML]{4BB05C}} \color[HTML]{F1F1F1} 0.437 & {\cellcolor[HTML]{006837}} \color[HTML]{F1F1F1} 0.594 \\
BMC-CLIP & {\cellcolor[HTML]{0C7F43}} \color[HTML]{F1F1F1} 0.701 & {\cellcolor[HTML]{C5E67E}} \color[HTML]{000000} 0.546 & {\cellcolor[HTML]{026C39}} \color[HTML]{F1F1F1} 0.788 & {\cellcolor[HTML]{199750}} \color[HTML]{F1F1F1} 0.424 & {\cellcolor[HTML]{0C7F43}} \color[HTML]{F1F1F1} 0.824 & {\cellcolor[HTML]{CDEA83}} \color[HTML]{000000} 0.648 & {\cellcolor[HTML]{8CCD67}} \color[HTML]{000000} 0.616 & {\cellcolor[HTML]{EF633F}} \color[HTML]{F1F1F1} 0.088 & {\cellcolor[HTML]{04703B}} \color[HTML]{F1F1F1} 0.745 & {\cellcolor[HTML]{57B65F}} \color[HTML]{F1F1F1} 0.436 & {\cellcolor[HTML]{04703B}} \color[HTML]{F1F1F1} 0.592 \\
\midrule
MAE & {\cellcolor[HTML]{FDBB6C}} \color[HTML]{000000} 0.574 & {\cellcolor[HTML]{006837}} \color[HTML]{F1F1F1} 0.572 & {\cellcolor[HTML]{0F8446}} \color[HTML]{F1F1F1} 0.777 & {\cellcolor[HTML]{148E4B}} \color[HTML]{F1F1F1} 0.426 & {\cellcolor[HTML]{15904C}} \color[HTML]{F1F1F1} 0.816 & {\cellcolor[HTML]{118848}} \color[HTML]{F1F1F1} 0.728 & {\cellcolor[HTML]{006837}} \color[HTML]{F1F1F1} 0.639 & {\cellcolor[HTML]{B10B26}} \color[HTML]{F1F1F1} 0.070 & {\cellcolor[HTML]{B9E176}} \color[HTML]{000000} 0.695 & {\cellcolor[HTML]{006837}} \color[HTML]{F1F1F1} 0.447 & {\cellcolor[HTML]{6EC064}} \color[HTML]{000000} 0.572 \\
DINOv2 & {\cellcolor[HTML]{FFFEBE}} \color[HTML]{000000} 0.609 & {\cellcolor[HTML]{FEE18D}} \color[HTML]{000000} 0.529 & {\cellcolor[HTML]{70C164}} \color[HTML]{000000} 0.742 & {\cellcolor[HTML]{96D268}} \color[HTML]{000000} 0.406 & {\cellcolor[HTML]{A9DA6C}} \color[HTML]{000000} 0.762 & {\cellcolor[HTML]{FDB163}} \color[HTML]{000000} 0.568 & {\cellcolor[HTML]{118848}} \color[HTML]{F1F1F1} 0.633 & {\cellcolor[HTML]{C21C27}} \color[HTML]{F1F1F1} 0.074 & {\cellcolor[HTML]{DDF191}} \color[HTML]{000000} 0.684 & {\cellcolor[HTML]{FFFEBE}} \color[HTML]{000000} 0.417 & {\cellcolor[HTML]{DCF08F}} \color[HTML]{000000} 0.552 \\
RAD-DINO & {\cellcolor[HTML]{279F53}} \color[HTML]{F1F1F1} 0.687 & {\cellcolor[HTML]{B91326}} \color[HTML]{F1F1F1} 0.503 & {\cellcolor[HTML]{E0F295}} \color[HTML]{000000} 0.697 & {\cellcolor[HTML]{70C164}} \color[HTML]{000000} 0.412 & {\cellcolor[HTML]{82C966}} \color[HTML]{000000} 0.777 & {\cellcolor[HTML]{FDB163}} \color[HTML]{000000} 0.568 & {\cellcolor[HTML]{FFF0A6}} \color[HTML]{000000} 0.590 & {\cellcolor[HTML]{B5DF74}} \color[HTML]{000000} 0.142 & {\cellcolor[HTML]{91D068}} \color[HTML]{000000} 0.706 & {\cellcolor[HTML]{F2FAAE}} \color[HTML]{000000} 0.419 & {\cellcolor[HTML]{A0D669}} \color[HTML]{000000} 0.564 \\
\midrule
\multicolumn{12}{c}{\textbf{Specialist}} \\
\midrule
CVNetGlobal50 & 0.695 & 0.644 & 0.814 & 0.455 & 0.937 & 0.793 & 0.686 & 0.090 & 0.868 & 0.503 & 0.644 \\
CVNetGlobal101 & 0.689 & 0.641 & 0.813 & 0.471 & 0.948 & 0.861 & 0.709 & 0.096 & 0.874 & 0.513 & 0.650 \\
\bottomrule
\end{tabular}
\end{table*}

\begin{table*}[t!]
\centering
    \caption{Retrieval performance of foundation models on the combined dataset, grouped by dataset.}
    \label{tab:retrieval_by_ds}
\begin{tabular}{c|cccc|cccc}
\toprule
& \multicolumn{4}{c|}{\textbf{micro P@1}} & \multicolumn{4}{c}{\textbf{macro P@1}}  \\ 
& \textbf{RadImageNet}    & \textbf{NIH14} & \textbf{MIMIC} & \textbf{CheXpert} &      \textbf{RadImageNet}    & \textbf{NIH14} & \textbf{MIMIC} & \textbf{CheXpert}      \\
& \textit{CT, MR, US} & \textit{XR} & \textit{XR} & \textit{XR} & \textit{CT, MR, US} & \textit{XR} & \textit{XR} & \textit{XR} \\
\midrule
\multicolumn{9}{c}{\textbf{Off-the-shelf}}          \\
\midrule
ResNet & {\cellcolor[HTML]{FFF7B2}} \color[HTML]{000000} 0.562 & {\cellcolor[HTML]{F7814C}} \color[HTML]{F1F1F1} 0.382 & {\cellcolor[HTML]{D22B27}} \color[HTML]{F1F1F1} 0.284 & {\cellcolor[HTML]{FEFFBE}} \color[HTML]{000000} 0.463 & {\cellcolor[HTML]{DDF191}} \color[HTML]{000000} 0.219 & {\cellcolor[HTML]{CC2627}} \color[HTML]{F1F1F1} 0.082 & {\cellcolor[HTML]{CE2827}} \color[HTML]{F1F1F1} 0.099 & {\cellcolor[HTML]{F7814C}} \color[HTML]{F1F1F1} 0.220 \\
ViT & {\cellcolor[HTML]{B5DF74}} \color[HTML]{000000} 0.585 & {\cellcolor[HTML]{E24731}} \color[HTML]{F1F1F1} 0.380 & {\cellcolor[HTML]{FDB163}} \color[HTML]{000000} 0.303 & {\cellcolor[HTML]{E0F295}} \color[HTML]{000000} 0.481 & {\cellcolor[HTML]{82C966}} \color[HTML]{000000} 0.235 & {\cellcolor[HTML]{B10B26}} \color[HTML]{F1F1F1} 0.078 & {\cellcolor[HTML]{E44C34}} \color[HTML]{F1F1F1} 0.105 & {\cellcolor[HTML]{DFF293}} \color[HTML]{000000} 0.314 \\
Ark & {\cellcolor[HTML]{199750}} \color[HTML]{F1F1F1} 0.612 & {\cellcolor[HTML]{006837}} \color[HTML]{F1F1F1} 0.398 & {\cellcolor[HTML]{006837}} \color[HTML]{F1F1F1} 0.364 & {\cellcolor[HTML]{4BB05C}} \color[HTML]{F1F1F1} 0.537 & {\cellcolor[HTML]{15904C}} \color[HTML]{F1F1F1} 0.250 & {\cellcolor[HTML]{006837}} \color[HTML]{F1F1F1} 0.149 & {\cellcolor[HTML]{026C39}} \color[HTML]{F1F1F1} 0.186 & {\cellcolor[HTML]{33A456}} \color[HTML]{F1F1F1} 0.389 \\
SAM & {\cellcolor[HTML]{FDB163}} \color[HTML]{000000} 0.542 & {\cellcolor[HTML]{E8F59F}} \color[HTML]{000000} 0.389 & {\cellcolor[HTML]{A50026}} \color[HTML]{F1F1F1} 0.276 & {\cellcolor[HTML]{FEE593}} \color[HTML]{000000} 0.444 & {\cellcolor[HTML]{FFF7B2}} \color[HTML]{000000} 0.208 & {\cellcolor[HTML]{BE1827}} \color[HTML]{F1F1F1} 0.080 & {\cellcolor[HTML]{C41E27}} \color[HTML]{F1F1F1} 0.097 & {\cellcolor[HTML]{A50026}} \color[HTML]{F1F1F1} 0.158 \\
MedSAM & {\cellcolor[HTML]{A50026}} \color[HTML]{F1F1F1} 0.506 & {\cellcolor[HTML]{FED27F}} \color[HTML]{000000} 0.385 & {\cellcolor[HTML]{F88C51}} \color[HTML]{F1F1F1} 0.298 & {\cellcolor[HTML]{A50026}} \color[HTML]{F1F1F1} 0.352 & {\cellcolor[HTML]{A50026}} \color[HTML]{F1F1F1} 0.163 & {\cellcolor[HTML]{B10B26}} \color[HTML]{F1F1F1} 0.078 & {\cellcolor[HTML]{DB382B}} \color[HTML]{F1F1F1} 0.102 & {\cellcolor[HTML]{D42D27}} \color[HTML]{F1F1F1} 0.183 \\
\midrule
CLIP & {\cellcolor[HTML]{78C565}} \color[HTML]{000000} 0.597 & {\cellcolor[HTML]{FFF5AE}} \color[HTML]{000000} 0.387 & {\cellcolor[HTML]{D22B27}} \color[HTML]{F1F1F1} 0.284 & {\cellcolor[HTML]{FEFFBE}} \color[HTML]{000000} 0.463 & {\cellcolor[HTML]{5DB961}} \color[HTML]{F1F1F1} 0.240 & {\cellcolor[HTML]{B10B26}} \color[HTML]{F1F1F1} 0.078 & {\cellcolor[HTML]{A90426}} \color[HTML]{F1F1F1} 0.092 & {\cellcolor[HTML]{F16640}} \color[HTML]{F1F1F1} 0.208 \\
MedCLIP & {\cellcolor[HTML]{C7E77F}} \color[HTML]{000000} 0.581 & {\cellcolor[HTML]{BB1526}} \color[HTML]{F1F1F1} 0.378 & {\cellcolor[HTML]{006837}} \color[HTML]{F1F1F1} 0.364 & {\cellcolor[HTML]{4BB05C}} \color[HTML]{F1F1F1} 0.537 & {\cellcolor[HTML]{D9EF8B}} \color[HTML]{000000} 0.220 & {\cellcolor[HTML]{138C4A}} \color[HTML]{F1F1F1} 0.143 & {\cellcolor[HTML]{006837}} \color[HTML]{F1F1F1} 0.187 & {\cellcolor[HTML]{BBE278}} \color[HTML]{000000} 0.334 \\
BiomedCLIP & {\cellcolor[HTML]{006837}} \color[HTML]{F1F1F1} 0.624 & {\cellcolor[HTML]{A50026}} \color[HTML]{F1F1F1} 0.377 & {\cellcolor[HTML]{FEDE89}} \color[HTML]{000000} 0.311 & {\cellcolor[HTML]{148E4B}} \color[HTML]{F1F1F1} 0.556 & {\cellcolor[HTML]{006837}} \color[HTML]{F1F1F1} 0.258 & {\cellcolor[HTML]{FDC574}} \color[HTML]{000000} 0.101 & {\cellcolor[HTML]{FDB96A}} \color[HTML]{000000} 0.122 & {\cellcolor[HTML]{006837}} \color[HTML]{F1F1F1} 0.425 \\
BMCA-CLIP & {\cellcolor[HTML]{04703B}} \color[HTML]{F1F1F1} 0.622 & {\cellcolor[HTML]{D22B27}} \color[HTML]{F1F1F1} 0.379 & {\cellcolor[HTML]{FDB163}} \color[HTML]{000000} 0.303 & {\cellcolor[HTML]{006837}} \color[HTML]{F1F1F1} 0.574 & {\cellcolor[HTML]{108647}} \color[HTML]{F1F1F1} 0.252 & {\cellcolor[HTML]{DD3D2D}} \color[HTML]{F1F1F1} 0.085 & {\cellcolor[HTML]{EA5739}} \color[HTML]{F1F1F1} 0.107 & {\cellcolor[HTML]{FFFEBE}} \color[HTML]{000000} 0.291 \\
\midrule
MAE & {\cellcolor[HTML]{6EC064}} \color[HTML]{000000} 0.599 & {\cellcolor[HTML]{FDB768}} \color[HTML]{000000} 0.384 & {\cellcolor[HTML]{FA9B58}} \color[HTML]{000000} 0.300 & {\cellcolor[HTML]{FEE593}} \color[HTML]{000000} 0.444 & {\cellcolor[HTML]{006837}} \color[HTML]{F1F1F1} 0.258 & {\cellcolor[HTML]{A50026}} \color[HTML]{F1F1F1} 0.076 & {\cellcolor[HTML]{E14430}} \color[HTML]{F1F1F1} 0.104 & {\cellcolor[HTML]{A50026}} \color[HTML]{F1F1F1} 0.158 \\
DINOv2 & {\cellcolor[HTML]{DCF08F}} \color[HTML]{000000} 0.576 & {\cellcolor[HTML]{FED27F}} \color[HTML]{000000} 0.385 & {\cellcolor[HTML]{C01A27}} \color[HTML]{F1F1F1} 0.281 & {\cellcolor[HTML]{E0F295}} \color[HTML]{000000} 0.481 & {\cellcolor[HTML]{82C966}} \color[HTML]{000000} 0.235 & {\cellcolor[HTML]{BE1827}} \color[HTML]{F1F1F1} 0.080 & {\cellcolor[HTML]{A50026}} \color[HTML]{F1F1F1} 0.091 & {\cellcolor[HTML]{F99355}} \color[HTML]{000000} 0.227 \\
RAD-DINO & {\cellcolor[HTML]{A2D76A}} \color[HTML]{000000} 0.589 & {\cellcolor[HTML]{FFF5AE}} \color[HTML]{000000} 0.387 & {\cellcolor[HTML]{BBE278}} \color[HTML]{000000} 0.334 & {\cellcolor[HTML]{006837}} \color[HTML]{F1F1F1} 0.574 & {\cellcolor[HTML]{66BD63}} \color[HTML]{F1F1F1} 0.239 & {\cellcolor[HTML]{9BD469}} \color[HTML]{000000} 0.128 & {\cellcolor[HTML]{BBE278}} \color[HTML]{000000} 0.154 & {\cellcolor[HTML]{A0D669}} \color[HTML]{000000} 0.347 \\
\midrule
\multicolumn{9}{c}{\textbf{Specialist}}          \\
\midrule
CVNetGlobal50 & 0.675 & 0.415 & 0.359 & 0.593 & 0.272 & 0.086 & 0.122 & 0.310 \\
CVNetGlobal101 & 0.681 & 0.430 & 0.371 & 0.556 & 0.287 & 0.092 & 0.137 & 0.253 \\
\bottomrule
\end{tabular}
\end{table*}

\end{document}